\documentclass{article}
\pdfoutput=1

\usepackage[square,sort,comma,numbers]{natbib}

\usepackage[preprint]{arxiv}

\usepackage[utf8]{inputenc} 
\usepackage[T1]{fontenc}    
\usepackage{hyperref}       
\usepackage{url}            
\usepackage{booktabs}       
\usepackage{amsfonts}       
\usepackage{nicefrac}       
\usepackage{microtype}      
\usepackage{lipsum}

\usepackage{xcolor}
\definecolor{codeblue}{rgb}{0,0.3,0.6}
\definecolor{codecrimson}{rgb}{0.678,0.063,0.184}  
\definecolor{codegreen}{rgb}{0,0.6,0}
\definecolor{codegray}{rgb}{0.5,0.5,0.5}
\definecolor{codepurple}{rgb}{0.58,0,0.82}
\definecolor{backcolour}{rgb}{0.95,0.95,0.92}

\usepackage{graphicx}
\usepackage{caption}
\usepackage{subcaption}

\usepackage{hyperref}       

\title{Technology Readiness Levels\\for Machine Learning Systems}

\author{
    \textbf{Alexander Lavin}\thanks{\url{lavin@simulation.science}} \\
    Pasteur Labs
\and
    \textbf{Ciar\'an M. Gilligan-Lee} \\
    Spotify
\and
    \textbf{Alessya Visnjic} \\
    WhyLabs
\and
    \textbf{Siddha Ganju} \\ 
    Nvidia
\\
\and
    \textbf{Dava Newman} \\
    MIT
\and
    \textbf{Atılım Güneş Baydin} \\
    University of Oxford
\and
    \textbf{Sujoy Ganguly} \\
    Unity AI
\and
    \textbf{Danny Lange} \\
    Unity AI
\\
\and
    \textbf{Amit Sharma}\\
    Microsoft Research
\and
    \textbf{Stephan Zheng}\\
    Salesforce Research
\and
    \textbf{Eric P. Xing}\\
    Petuum
\and
    \textbf{Adam Gibson}\\
    Konduit
\\
\and
    \textbf{James Parr}\\
    NASA Frontier Development Lab
\and
    \textbf{Chris Mattmann}\\
    NASA Jet Propulsion Lab
\and
    \textbf{Yarin Gal}\\
    Alan Turing Institute
}

\begin{document}
\maketitle

\begin{abstract}
The development and deployment of machine learning (ML) systems can be executed easily with modern tools, but the process is typically rushed and means-to-an-end. The lack of diligence can lead to technical debt, scope creep and misaligned objectives, model misuse and failures, and expensive consequences. Engineering systems, on the other hand, follow well-defined processes and testing standards to streamline development for high-quality, reliable results. The extreme is spacecraft systems, where mission critical measures and robustness are ingrained in the development process. Drawing on experience in both spacecraft engineering and ML (from research through product across domain areas), we have developed a proven systems engineering approach for machine learning development and deployment. Our \textit{Machine Learning Technology Readiness Levels (MLTRL)} framework defines a principled process to ensure robust, reliable, and responsible systems while being streamlined for ML workflows, including key distinctions from traditional software engineering. Even more, MLTRL defines a lingua franca for people across teams and organizations to work collaboratively on artificial intelligence and machine learning technologies. Here we describe the framework and elucidate it with several real world use-cases of developing ML methods from basic research through productization and deployment, in areas such as medical diagnostics, consumer computer vision, satellite imagery, and particle physics.
\end{abstract}


\renewcommand{\thefootnote}{\roman{footnote}}

\section*{Introduction}\label{sec_intro}

The accelerating use of artificial intelligence (AI) and machine learning (ML) technologies in systems of software, hardware, data, and people introduces vulnerabilities and risks due to dynamic and unreliable behaviors; fundamentally, ML systems learn from data, introducing known and unknown challenges in how these systems behave and interact with their environment.
Currently the approach to building AI technologies is siloed: models and algorithms are developed in testbeds isolated from real-world environments, and without the context of larger systems or broader products they'll be integrated within for deployment.
A main concern is models are typically trained and tested on only a handful of curated datasets, without measures and safeguards for future scenarios, and oblivious of the downstream tasks and users. 
Even more, models and algorithms are often integrated into a software stack without regard for the inherent stochasticity --for instance, the massive effect random seeds have on deep reinforcement learning model performance \cite{Henderson2018DeepRL} -- and failure modes of the ML components, which can be dangerously hidden in layers of software and abstraction.


Other domains of engineering, such as civil and aerospace, follow well-defined processes and testing standards to streamline development for high-quality, reliable results. \textit{Technology Readiness Level (TRL)} is a systems engineering protocol for deep tech\cite{deeptech} and scientific endeavors at scale, ideal for integrating many interdependent components \textit{and} cross-functional teams of people. It is no surprise that TRL is standard process and parlance in NASA\cite{Nasa2003NASASE} and DARPA\cite{dod}.

For a spaceflight project there are several defined phases, from pre-concept to prototyping to deployed operations to end-of-life, each with a series of exacting development cycles and reviews. This is in stark contrast to common machine learning and software workflows, which promote quick iteration, rapid deployment, and simple linear progressions. Yet the NASA technology readiness process for spacecraft systems is overkill;
we need robust ML technologies integrated with larger systems of software, hardware, data, and humans, but not necessarily for missions to Mars. 
We aim to bring systems engineering to AI and ML by defining and putting into action a lean \textit{Machine Learning Technology Readiness Levels (MLTRL)} framework. We draw on decades of AI and ML development, from research through production, across domains and diverse data scenarios: for example, computer vision in medical diagnostics and consumer apps, automation in self-driving vehicles and factory robotics, tools for scientific discovery and causal inference, streaming time-series in predictive maintenance and finance.

In this paper we define our framework for developing and deploying robust, reliable, and responsible ML and data systems, with several real test cases of advancing models and algorithms from R\&D through productization and deployment, including essential data considerations.
Additionally, MLTRL prioritizes the role of AI ethics and fairness, and our systems AI approach can help curb the large societal issues that can result from poorly deployed and maintained AI and ML technologies, such as the automation of systemic human bias, denial of individual autonomy, and unjustifiable outcomes (see the \textit{Alan Turing Institute Report on Ethical AI}\cite{Leslie2019UnderstandingAI}).
The adoption and proliferation of MLTRL provides a common nomenclature and metric across teams and industries.
The standardization of MLTRL across the AI industry should help teams and organizations develop principled, safe, and trusted technologies.

\begin{figure}[ht]
\centering
\includegraphics[width=\linewidth]{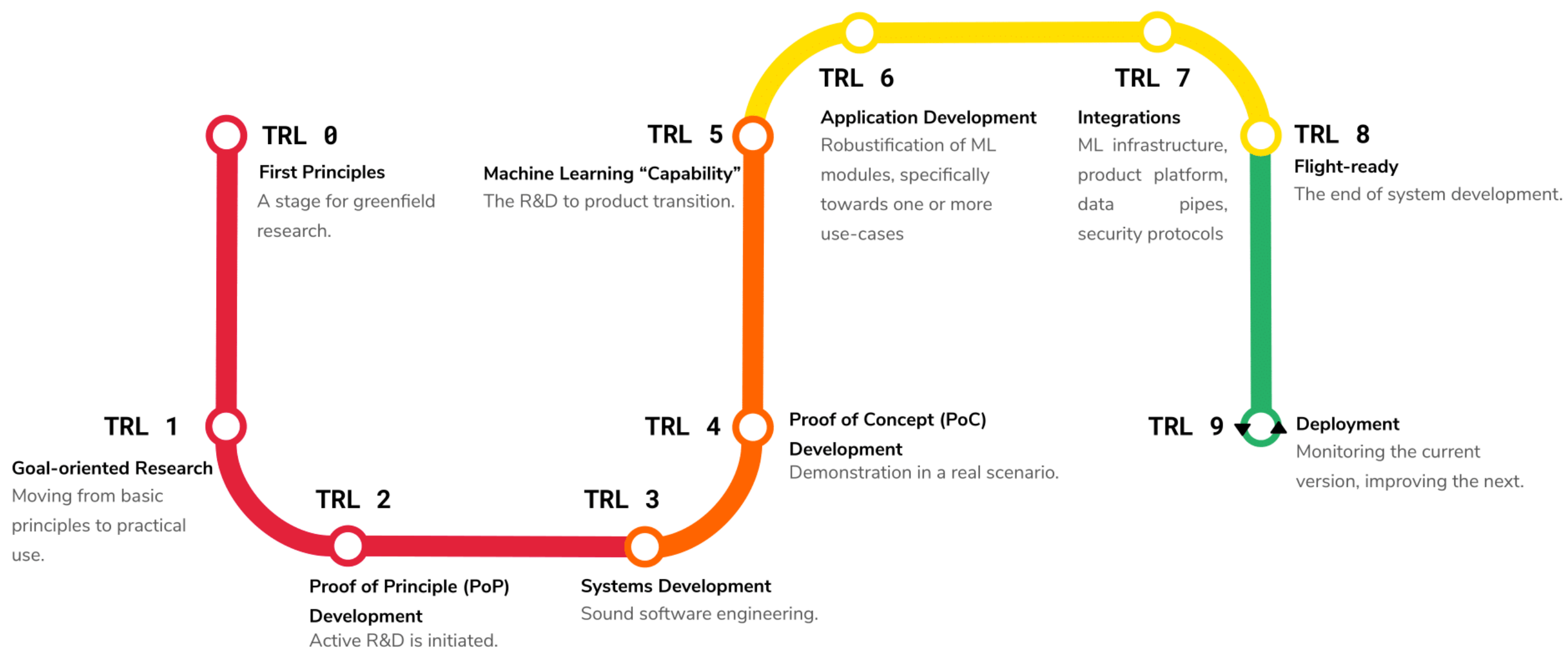}
\caption{
MLTRL spans research through prototyping, productization, and deployment. Most ML workflows prescribe an isolated, linear process of data processing, training, testing, and serving a model~\cite{mlworkflow}. Those workflows fail to define how ML development must iterate over that basic process to become more mature and robust, and how to integrate with a much larger system of software, hardware, data, and people.
Not to mention MLTRL continues beyond deployment: monitoring and feedback cycles are important for continuous reliability and improvement over the product lifetime.
}
\label{fig:trl4ml}
\end{figure}

\section*{Results}

\textit{MLTRL} defines technology readiness levels (TRLs) to guide and communicate AI and ML development and deployment. A TRL represents the maturity of a model or algorithm\footnote{Note we use ``model'' and ``algorithm'' somewhat interchangeably when referring to the technology under development. The same MLTRL process and methods apply for a machine translation model and for an A/B testing algorithm, for example.}, data pipelines, software module, or composition thereof; a typical ML system consists of many interconnected subsystems and components, and the TRL of the system is the lowest level of its constituent parts~\cite{Lavin2020TechnologyRL}. The anatomy of a level is marked by gated reviews, evolving working groups, requirements documentation with risk calculations, progressive code and testing standards, and deliverables such as TRL Cards (Figure \ref{fig:trl_card}) and ethics checklists.\footnote{Templates and examples for MLTRL deliverables will be open-sourced upon publication at \href{https://github.com/alan-turing-institute}{github.com/alan-turing-institute}.} These components---which are crucial for implementing the levels in a systematic fashion---as well as MLTRL metrics and methods are concretely described in examples and in the Methods section. Lastly, to emphasize the importance of data tasks in ML, from data curation~\cite{Dasu2003ExploratoryDM} to data governance~\cite{Janssen2020DataGO}, we state several important data considerations at each MLTRL level. 

\subsection*{MACHINE LEARNING TECHNOLOGY READINESS LEVELS}\label{sec_levels}
The levels are briefly defined as follows and in Figure \ref{fig:trl4ml}, and elucidated with real-world examples later.

\begin{figure}[t]
\centering
\includegraphics[width=\linewidth]{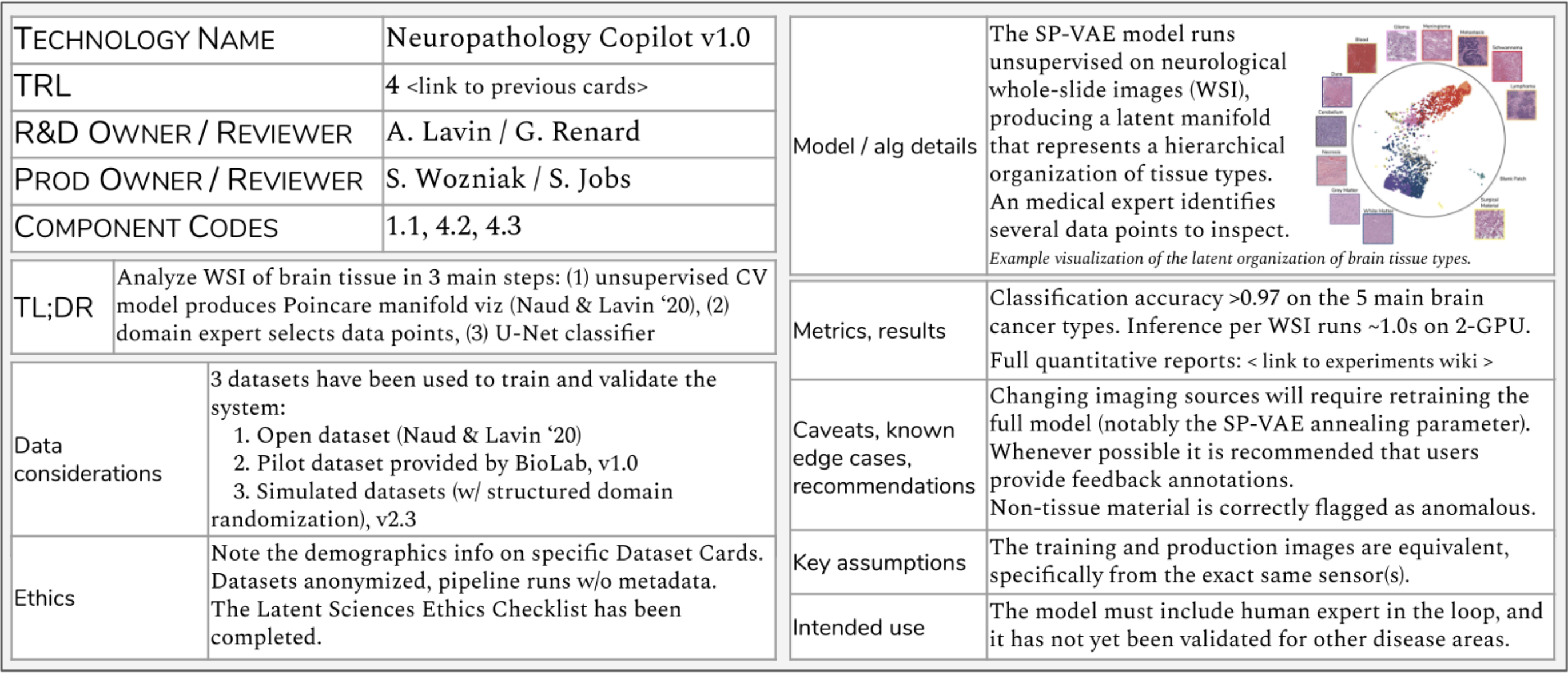}
\caption{
The maturity of each ML technology is tracked via \textit{TRL Cards}, which we describe in the Methods section. Here is an example reflecting a neuropathology machine vision use-case\cite{Naud2020ManifoldsFU}, detailed in the Discussion Section. Note this is a subset of a full TRL Card, which in reality lives as a full document in an internal wiki.
Notice the card clearly communicates the data sources, versions, and assumptions. This helps mitigate invalid assumptions about performance and generalizability when moving from R\&D to production, and promotes the use of real-world data earlier in the project lifecycle. We recommend documenting datasets thoroughly with semantic versioning and tools such as \textit{datasheets for datasets}\cite{Gebru2018DatasheetsFD}, and following data accountability best-practices as they evolve (see \cite{Hutchinson2021TowardsAF}).
}
\label{fig:trl_card}
\end{figure}

\paragraph{Level 0 - First Principles}
This is a stage for greenfield AI research, initiated with a novel idea, guiding question, or poking at a problem from new angles.
The work mainly consists of literature review, building mathematical foundations, white-boarding concepts and algorithms, and building an understanding of the data -- for work in theoretical AI and ML, however, there will not yet be data to work with (for example, a novel algorithm for Bayesian optimization\cite{Shahriari2016TakingTH}, which could eventually be used for many domains and datasets).
The outcome of Level 0 is a set of concrete ideas with sound mathematical formulation, to pursue through low-level experimentation in the next stage. When relevant, this level expects conclusions about data readiness, including strategies for getting the data to be suitable for the specific ML task.
To graduate, the basic principles, hypotheses, data readiness, and research plans need to be stated, referencing relevant literature. 
With graduation, a \textit{TRL Card} should be started to succinctly document the methods and insights thus far -- this key MLTRL deliverable is detailed in the Methods section and Figure \ref{fig:trl_card}.

\textit{Level 0 data -- }
Not a hard requirement at this stage  because this is largely theoretical machine learning. That being said, data availability needs to be considered for defining any research project to move past theory.

\textit{Level 0 review --}
The reviewer here is solely the lead of the research lab or team, for instance a PhD supervisor. We assess hypotheses and explorations for mathematical validity and potential novelty or utility, not necessarily code nor end-to-end experiment results.

\paragraph{Level 1 - Goal-Oriented Research} 
To progress from basic principles to practical use, we design and run low-level experiments to analyze specific model or algorithm properties (rather than end-to-end runs for a performance benchmark score).
This involves collection and processing of sample data to train and evaluate the model. This sample data need not be the full data; it may be a smaller sample that is currently available or more convenient to collect. In some cases it may suffice to use synthetic data as the representative sample -- in the medical domain, for example, acquiring datasets can take many months due to security and privacy constraints, so generating sample data can mitigate this blocker from early ML development.
Further, working with the sample data provides a blueprint for the data collection and processing pipeline (including answering  whether it is even possible to collect all necessary data), that can be scaled up for the  for the next steps.
The experiments, good results or not, and mathematical foundations need to pass a review process with fellow researchers before graduating to Level 2.
The application is still speculative, but through comparison studies and analyses we start to understand if/how/where the technology offers potential improvements and utility. 
Code is \textit{research-caliber}: The aim here is to be quick and dirty, moving fast through iterations of experiments. Hacky code is okay, and full test coverage is actually discouraged, as long as the overall codebase is organized and maintainable. 
It is important to start semantic versioning practices early in the project lifecycle, which should cover code, models, \textit{and} datasets. This is crucial for retrospectives and reproducibility, issues with which can be costly and severe at later stages.
This versioning information and additional progress should be reported on the TRL Card (see for example Figure \ref{fig:trl_card}).

\textit{Level 1 data -- }
At minimum we work with sample data that is representative of downstream real datasets, which can be a subset of real data, synthetic data, or both. Beyond driving low-level ML experiments, the sample data forces us to consider data acquisition and processing strategies at an early stage before it becomes a blocker later.

\textit{Level 1 review --}
The panel for this gated review is entirely members of the research team, reviewing for scientific rigor in early experimentation, and pointing to important concepts and prior work from their respective areas of expertise. There may be several iterations of feedback and additional experiments.

\paragraph{Level 2 - Proof of Principle (PoP) Development} 
Active R\&D is initiated, mainly by developing and running in \textit{testbeds}: simulated environments and/or simulated data that closely matches the conditions and data of real scenarios -- note these are driven by model-specific technical goals, not necessarily application or product goals (yet). 
An important deliverable at this stage is the formal research requirements document (with well-specified verification and validation (V\&V) steps)\footnote{A \textit{requirement} is a singular documented physical or functional need that a particular design, product, or process aims to satisfy. Requirements aim to specify all stakeholders' needs while not specifying a specific solution. Definitions are incomplete without corresponding measures for verification and validation (V\&V). \textit{Verification: Are we building the product right?} \textit{Validation: Are we building the right product?}}.
Here is one of several \textit{key decision points} in the broader process: The R\&D team considers several paths forward and sets the course: (A) prototype development towards Level 3, (B) continued R\&D for longer-term research initiatives and/or publications, or some combination of A and B. 
We find the culmination of this stage is often a bifurcation: some work moves to applied AI, while some circles back for more research. This common MLTRL cycle is an instance of the non-monotonic \textit{discovery switchback} mechanism (detailed in the Methods section).

\textit{Level 2 data -- }
Datasets at this stage may include publicly available benchmark datasets, semi-simulated data based on the data sample in Level 1, or fully simulated data based on certain assumptions about the potential deployment environments. The data should allow researchers to characterize model properties, and highlight corner cases or boundary conditions, in order to justify the utility of continuing R\&D on the model. 

\textit{Level 2 review --}
To graduate from the PoP stage, the technology needs to satisfy research claims made in previous stages (brought to be bare by the aforementioned PoP data in both quantitative and qualitative ways) with the analyses well-documented and reproducible.

\paragraph{Level 3 - System Development} 
Here we have checkpoints that push code development towards interoperability, reliability, maintainability, extensibility, and scalability. 
Code becomes \textit{prototype-caliber}: A significant step up from research code in robustness and cleanliness. This needs to be well-designed, well-architected for dataflow and interfaces, generally covered by unit and integration tests, meet team style standards, and sufficiently-documented.
Note the programmers' mentality remains that this code will someday be refactored/scrapped for productization; prototype code is relatively primitive with regard to efficiency and reliability of the eventual system.
With the transition to Level 4 and proof-of-concept mode, the working group should evolve to include product engineering to help define service-level agreements and objectives (SLAs and SLOs) of the eventual production system.

\textit{Level 3 data -- }
For the most part consistent with Level 2; in general, the previous level review can elucidate potential gaps in data coverage and robustness to be addressed in the subsequent level. However, for test suites developed at this stage, it is useful to define dedicated subsets of the experiment data as default testing sources, as well as setup mock data for specific functionalities and scenarios to be tested.

\textit{Level 3 review --}
Teammates from applied AI and engineering are brought into the review to focus on sound software practices, interfaces and documentation for future development, and version control for models and datasets. There are likely domain- or organization-specific data management considerations going forward that this review should point out -- e.g. standards for data tracking and compliance in healthcare \cite{Ramakrishnan2020TowardsCD}.

\paragraph{Level 4 - Proof of Concept (PoC) Development}
This stage is the seed of application-driven development; for many organizations this is the first touch-point with product managers and stakeholders beyond the R\&D group. Thus TRL Cards and requirements documentation are instrumental in communicating the project status and onboarding new people.
The aim is to demonstrate the technology in a real scenario: quick proof-of-concept examples are developed to explore candidate application areas and communicate the quantitative and qualitative results.
It is essential to use real and representative data for these potential applications. Thus data engineering for the PoC largely involves scaling up the data collection and processing from Level 1, which may include collecting new data or processing all available data using scaled experiment pipelines from Level 3. In some scenarios there will new datasets brought in for the PoC, for example, from an external research partner as a means of validation.
Hand-in-hand with the evolution from sample to real data, the experiment metrics should evolve from ML research to the applied setting: proof-of-concept evaluations should quantify model and algorithm performance (e.g., precision and recall and various data splits), computational costs (e.g., CPU vs GPU runtimes), and also metrics that are more relevant to the eventual end-user (e.g., number of false positives in the top-N predictions of a recommender system).
We find this PoC exploration reveals specific differences between clean and controlled research data versus noisy and stochastic real-world data. The issues can be readily identified because of the well-defined distinctions between those development stages in MLTRL, and then targeted for further development.

AI ethics processes vary across organizations, but all should engage in ethics conversations at this stage, including ethics of data collection, and potential of any harm or discriminatory impacts due to the model (as the AI capabilities and datasets are known).
MLTRL requires ethics considerations to be reported on TRL Cards at all stages, which generally link to an extended ethics checklist.
The \textit{key decision point} here is to push onward with application development or not. It is common to pause projects that pass Level 4 review, waiting for a better time to dedicate resources, and/or pull the technology into a different project.

\textit{Level 4 data -- }
Unlike the previous stages, having real-world and representative  data is critical for the PoC; even with methods for verifying that data distributions in synthetic data reliably mirror those of real data \cite{}, sufficient confidence in the technology must be achieved with real-world data of the use-case.
Further, one must consider how to obtain high-quality and consistent data required for the future model inference: generation of the data pipeline PoC that will resemble the future inference pipeline that will take data from intended sources, transform it into features, and send it to the model for inference.

\textit{Level 4 review --}
Demonstrate the utility towards one or more practical applications (each with multiple datasets), taking care to communicate assumptions and limitations, and again reviewing data-readiness: evaluating the real-world data for quality, validity, and availability. The review also evaluates security and privacy considerations -- defining these in the requirements document with risk quantification is a useful mechanism for mitigating potential issues (discussed further in the Methods section).

\begin{figure}[ht]
\centering
\includegraphics[width=0.85\linewidth]{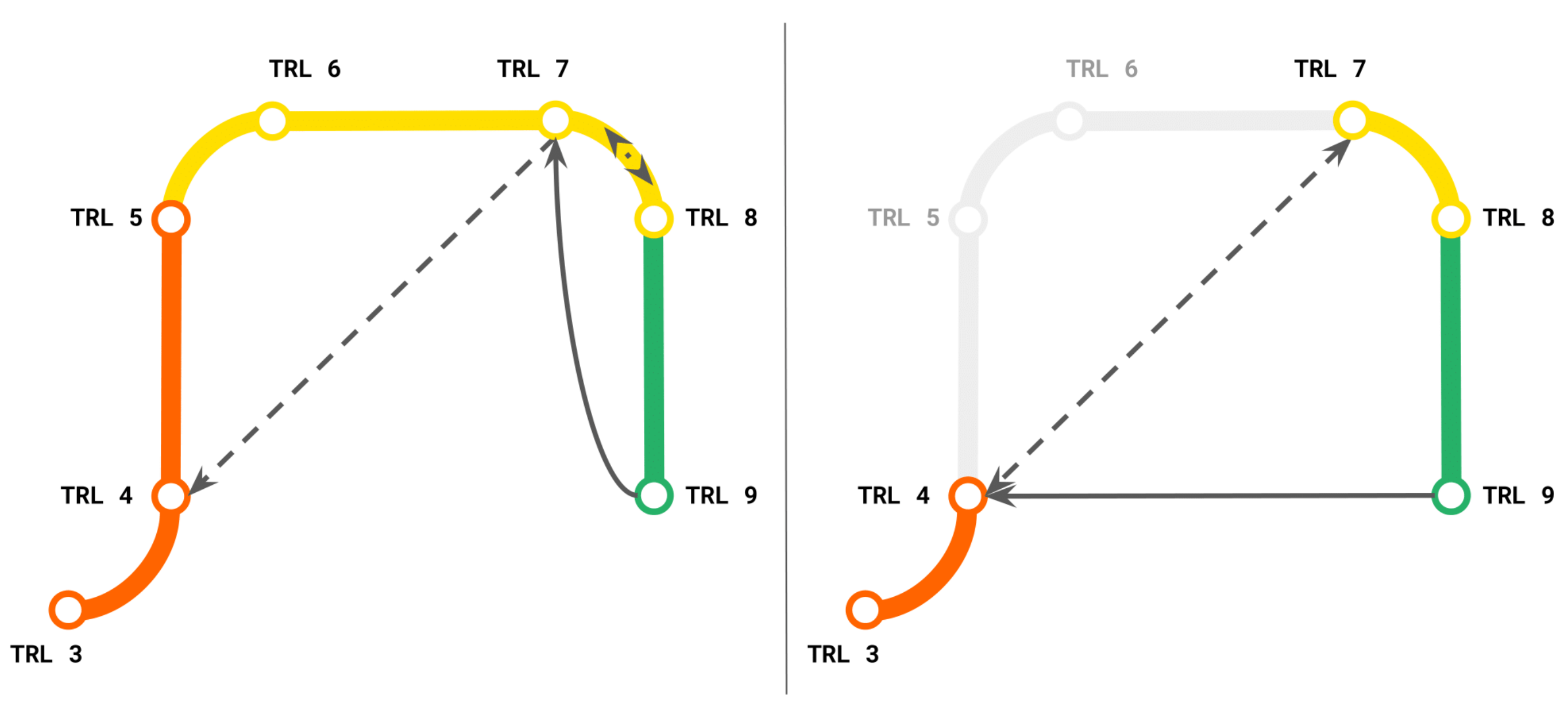}
\caption{
Most ML and AI projects live in these sections of MLTRL, not concerned with fundamental R\&D -- that is, completely using existing methods and implementations, and even pretrained models.
In the left diagram, the arrows show a common development pattern with MLTRL in industry: projects go back to the ML toolbox to develop new features (dashed line), and frequent, incremental improvements are often a practice of jumping back a couple levels to Level 7 (which is the main systems integrations stage). At Levels 7 and 8 we stress the need for tests that run use-case specific critical scenarios and data-slices, which are highlighted by a proper risk-quantification matrix~\cite{Duijm2015RecommendationsOT}.
Cycling back to previous lower levels is not just a late-stage mechanism in MLTRL, but rather ``switchbacks'' occur throughout the process (as discussed in the Methods section and throughout the text).
In the right diagram we show the more common approach in industry (\textit{without} using our framework), which skips essential technology transition stages -- ML Engineers push straight through to deployment, ignoring important productization and systems integration factors.
This will be discussed in more detail in the Methods section.
}
\label{fig:trl_flows}
\end{figure}

\paragraph{Level 5 - Machine Learning ``Capability''} 
At this stage the technology is more than an isolated model or algorithm, it is a specific \textit{capability}. For instance, producing depth images from stereo vision sensors on a mobile robot is a real-world capability beyond the isolated ML technique of self-supervised learning for RGB stereo disparity estimation.
In many organizations this represents a technology transition or handoff from R\&D to productization. MLTRL makes this transition explicit, evolving the requisite work, guiding documentation, objectives and metrics, and team; indeed, without MLTRL it is common for this stage to be erroneously leaped completely, as shown in Figure \ref{fig:trl_flows}.
An interdisciplinary working group is defined, as we start developing the technology in the context of a larger real-world process – i.e., transitioning the model or algorithm from an isolated solution to a module of a larger application.
Just as the ML technology should no longer be owned entirely by ML experts, steps have been taken to share the technology with others in the organization via demos, example scripts, and/or an API; the knowledge and expertise cannot remain within the R\&D team, let alone an individual ML developer. 
Graduation from Level 5 should be difficult, as it signifies the dedication of resources to push this ML technology through productization.
This transition is a common challenge in deep-tech, sometimes referred to as ``the valley of death'' because project managers and decision-makers struggle to allocate resources and align technology roadmaps to effectively move to Level 6, 7 and onward.
MLTRL directly addresses this challenge by stepping through the technology transition or handoff explicitly.

\textit{Level 5 data --}
For the most part consistent with Level 4. However, considerations need to be taken for scaling of data pipelines: there will soon be more engineers accessing the existing data and adding more, and the data will be getting much more use, including automated testing in later levels. With this scaling can come challenges with data governance. The data pipelines likely do not mirror the structure of the teams or broader organization. This can result in data silos, duplications, unclear responsibilities, and missing control of data over its entire lifecycle. These challenges and several approaches to data governance (planning and control, organizational, and risk-based) are detailed in Janssen et al. \cite{Janssen2020DataGO}.

\textit{Level 5 review --}
The verification and validation (V\&V) measures and steps defined in earlier R\&D stages (namely Level 2) must all be completed by now, and the product-driven requirements (and corresponding V\&V) are drafted at this stage. We thoroughly review them here, and make sure there is stakeholder alignment (at the first possible step of productization, well ahead of deployment).

\paragraph{Level 6 - Application Development} 
The main work here is significant software engineering to bring the code up to \textit{product-caliber}:
This code will be deployed to users and thus needs to follow precise specifications, have comprehensive test coverage, well-defined APIs, etc.
The resulting ML modules should be robustified towards one or more target use-cases. 
If those target use-cases call for model explanations, the methods need to be built and validated alongside the ML model, and tested for their efficacy in faithfully interpreting the model's decisions -- crucially, this needs to be in the context of downstream tasks and the end-users, as there is often a gap between ML explainability that serves ML engineers rather than external stakeholders\cite{Bhatt2020ExplainableML}.
Similarly, we need to develop the ML modules with known data challenges in mind, specifically to check the robustness of the model (and broader pipeline) to changes in the data distribution between development and  deployment. 

The deployment setting(s) should be addressed thoroughly in the product requirements document, as ML serving (or deploying) is an overloaded term that needs careful consideration.
First, there are two main types: internal, as APIs for experiments and other usage mainly by data science and ML teams, and external, meaning an ML model that is embedded or consumed within a real application with real users.
The serving constraints vary significantly when considering cloud deployment vs on-premise or hybrid, batch or streaming, open-source solution or containerized executable, etc. 
Even more, the data at deployment may be limited due to compliance, or we may only have access to encrypted data sources, some of which may only be accessible locally -- these scenarios may call for advanced ML approaches such as federated learning\cite{Li2020FederatedLC} and other privacy-oriented ML\cite{Ryffel2018AGF}.
And depending on the application, an ML model may not be deployable without restrictions; this typically means being embedded in a rules engine workflow where the ML model acts like an advisor that discovers edge cases in rules.
These deployment factors are hardly considered in model and algorithm development despite significant influence on modeling and algorithmic choices; that said, hardware choices typically are considered early on, such as GPU versus edge devices. It is crucial to make these systems decisions at Level 6--not too early that serving scenarios and requirements are uncertain, and not too late that corresponding changes to model or application development risk deployment delays or failures.
This marks a \textit{key decision} for the project lifecycle, as this expensive ML deployment risk is common without MLTRL (see Figure \ref{fig:trl_flows}).

\textit{Level 6 data --}
Additional data should be collected and operationalized at this stage towards robustifying the ML models, algorithms, and surrounding components.
These include adversarial examples to check local robustness~\cite{Madry2018TowardsDL}, semantically-equivalent perturbations to check consistency of the model with respect to domain assumptions~\cite{Zhao2018GeneratingNA, Ribeiro2020BeyondAB}, and collecting data from different sources and checking how well the trained model generalizes to them. These considerations are even more vital in the challenging deployment domains mentioned above with limited data access.

\textit{Level 6 review --}
Focus is on the code quality, the set of newly defined product requirements, system SLA and SLO requirements, data pipelines spec, and an AI ethics revisit now that we are closer to a real-world use-case. In particular, regulatory compliance is mandated for this gated review; the data privacy and security laws are changing rapidly, and missteps with compliance can make or break the project.

\paragraph{Level 7 - Integrations} 
For integrating the technology into existing production systems, we recommend the working group has a balance of infrastructure engineers \textit{and} applied AI engineers -- this stage of development is vulnerable to latent model assumptions and failure modes, and as such cannot be safely developed solely by software engineers.
Important tools for them to build together include:
\begin{itemize}
  \item Tests that run use-case specific critical scenarios and data-slices -- a proper risk-quantification table will highlight these.
  \item A ``golden dataset'' should be defined to baseline the performance of each model \textit{and succession of models}--see the computer vision app example in Figure \ref{fig:unity}--for use in the continuous integration and deployment (CI/CD) tests.
  \item \textit{Metamorphic testing}: a software engineering methodology for testing a specific set of relations between the outputs of multiple inputs. When integrating ML modules into larger systems, a codified list of metamorphic relations\cite{Xie2011TestingAV} can provide valuable verification and validation measures and steps.
  \item \textit{Data intervention tests} that seek data bugs at various points in the pipelines, downstream to measure the potential effects of data processing and ML on consumers or users of that data, as well as upstream at data ingestion or creation. Rather than using model performance as a proxy for data quality, it is crucial to use intervention tests that instead catch data errors with mechanisms specific to data validation.
\end{itemize}
These tests in particular help mitigate underspecification in ML pipelines, a key obstacle to reliably training models that behave as expected in deployment\cite{DAmour2020UnderspecificationPC}.
On the note of reliability, it is important that quality assurance engineers (QA) play a key role here and through Level 9, overseeing data processes to ensure privacy and security, and covering audits for downstream accountability of AI methods.


\textit{Level 7 data --}
In addition to the data for test suites discussed above, this level calls for QA to prioritize \textit{data governance}: how data is obtained, managed, used, and secured by the organization. This was earlier suggested in level 5 (in order to preempt related technical debt), and essential here at the main junction for integration, which may create additional governance challenges in light of downstream effects and consumers. 

\textit{Level 7 review --}
The review should focus on the data pipelines and test suites; a scorecard like the ML Testing Rubric\cite{Breck2017TheMT} is useful. The group should also emphasize ethical considerations at this stage, as they may be more adequately addressed now (where there are many test suites put into place) rather than close to shipping later.

\paragraph{Level 8 - Flight-ready} 
The technology is demonstrated to work in its final form and under expected conditions. There should be additional tests implemented at this stage covering deployment aspects, notably A/B tests, blue/green deployment tests, shadow testing, and canary testing, which enable proactive and gradual testing for changing ML methods and data. 
Ahead of deployment, the CI/CD system should be ready to regularly stress test the overall system and ML components.
In practice, problems stemming from real-world data are impossible to anticipate and design for -- an upstream data provider could change formats unexpectedly or a physical event could cause the customer behavior to change. Running models in shadow mode for a period of time would help stress test the infrastructure and evaluate how susceptible the ML model(s) will be to performance regressions caused by data. 
We observe that ML systems with \textit{data-oriented architectures} are more readily tested in this manner, and better surface data quality issues, data drifts, and concept drifts -- this is discussed later in the Beyond Software Engineering section.
To close this stage, the \textit{key decision} is go or no-go for deployment, and when.

\textit{Level 8 data --}
If not already in place, there absolutely needs to be mechanisms for automatically logging data distributions alongside model performance once deployed.

\textit{Level 8 review --}
A diligent walkthrough of every technical and product requirement, showing the corresponding validations, and the review panel is representative of the full slate of stakeholders.

\paragraph{Level 9 - Deployment} 
In deploying AI and ML technologies, there is significant need to monitor the current version, and explicit considerations towards improving the next version. For instance, performance degradation can be hidden and critical, and feature improvements often bring unintended consequences and constraints.
Thus at this level, the focus is on maintenance engineering--i.e., methods and pipelines for ML monitoring and updating.
Monitoring for data quality, concept drift, and data drift is crucial; no AI system without thorough tests for these can reliably be deployed. By the same token there must be automated evaluation and reporting -- if actuals\cite{Botchkarev2019ANT} are available, continuous evaluation should be enabled, but in many cases actuals come with a delay, so it is essential to record model outputs to allow  for efficient evaluation after the fact.
To these ends, the ML pipeline should be instrumented to log system metadata, model metadata, and data itself. 

Monitoring for data quality issues and data drifts is crucial to catch deviations in model behavior, particularly those that are non-obvious in the model or product end-performance. Data logging is unique in the context of ML systems: data logs should capture statistical properties of input features and model predictions, and capture their anomalies. With monitoring for data, concept, and model drifts, the logs are to be sent to the relevant systems, applied, and research engineers. The latter is often non-trivial, as the model server is not ideal for model ``observability'' because it does not necessarily have the right data points to link the complex layers needed to analyze and debug models. To this end, MLTRL requires the drift tests to be implemented at stages well ahead of deployment, earlier than is standard practice. Again we advocate for data-first architectures rather than the software industry-standard design by services (discussed later), which aids in surfacing and logging the relevant data types and slices when monitoring AI systems.
For retraining and improving models, monitoring must be enabled to catch training-serving skew and let the team know when to retrain. Towards model improvements, adding or modifying features can often have unintended consequences, such as introducing latencies or even bias. To mitigate these risks, MLTRL has an \textit{embedded switchback} here: any component or module changes to the deployed version \textit{must} cycle back to Level 7 (integrations stage) or earlier.
Additionally, for quality ML products, we stress a defined communication path for user feedback without roadblocks to R\&D; we encourage real-world feedback all the way to research, providing valuable problem constraints and perspectives.

\textit{Level 9 data --}
Proper mechanisms for logging and inspecting data (alongside models) is critical for deploying reliable AI and ML -- systems that learn on data have unique monitoring requirements (detailed above). In addition to the infrastructure and test suites covering data and environment shifts, it's important for product managers and other owners to be on top of data policy shifts in domains such as finance and healthcare.

\textit{Level 9 review --}
The review at this stage is unique, as it also helps in lifecycle management: at a regular cadence that depends on the deployed system and domain of use, owners and other stakeholders are to revisit this review and recommend switchbacks if needed (discussed in the Methods section). This additional oversight at deployment is shown to help define regimented release cycles of updated versions, and provide another ``eye'' check for stale model performance or other system abnormalities.

\medskip

\noindent
Notice MLTRL is defined as stages or levels, yet much of the value in practice is realized in the transitions: MLTRL enables teams to move from one level to the next reliably and efficiently, and provides a guide for how teams and objectives evolve with the progressing technology.


\section*{Discussion}\label{sec_disc}

MLTRL is designed to apply to many real-world use-cases involving data and ML, from simple regression models used for predictive modeling energy demand or anomaly detection in datacenters, to real-time modeling in rideshare applications and motion planning in warehouse robotics.
For simple use-cases MLTRL may be overkill, and a subset may suffice -- for instance, model cards as demonstrated by Google for basic image classification. Yet this is a fine line, as the same cards-only approach in the popular ``Huggingface'' codebases are too simplistic for the language models they represent, deployed in domains that carry significant consequences.
MLTRL becomes more valuable with more complex, larger systems and environments, especially in risk averse domains. We thoroughly discuss this through several real uses of MLTRL below.

\subsection*{EXAMPLES}\label{sec_examples}

\subsubsection*{Human-machine visual inspection}
While most ML projects begin with a specific task and/or dataset, there are many that originate in ML theory without any target application -- i.e., projects starting MLTRL at level 0 or 1. These projects nicely demonstrate the utility of MLTRL built-in switchbacks, bifurcating paths, and 
iteration with domain experts.
An example we discuss here is a novel approach to representing data in generative vision models from Naud \& Lavin\cite{Naud2020ManifoldsFU}, which was then developed into state-of-the-art unsupervised anomaly detection, and targeted for two human-machine visual inspection applications: 
First, industrial anomaly detection, notably in precision manufacturing, to identify potential errors for human-expert manual inspection.
Second, using the model to improve the accuracy and efficiency of neuropathology, the microscopic examination of neurosurgical specimens for cancerous tissue.
In these human-machine teaming use-cases there are specific challenges impeding practical, reliable use:



\begin{itemize}
    \item \textbf{Hidden feedback loops} can be common and problematic in real-world systems influencing their own training data: over time the behavior of users may evolve to select data inputs they prefer for the specific AI system, representing some skew from the training data. In this neuropathology case, selecting whole-slide images that are uniquely difficult for manual inspection, or even biased by that individual user.
    Similarly we see underlying healthcare processes can act as hidden confounders, resulting in unreliable decision support tools\cite{Schulam2017ReliableDS}.
    \item \textbf{Model availability} can be limited in many deployment settings: for example, on-premises deployments (common in privacy preserving domains like healthcare and banking), edge deployments (common in industrial use-cases such as manufacturing and agriculture), or from the infrastructure's inability to scale to the volume of requests. This can severely limit the team's ability to monitor, debug, and improve deployed models.
    \item \textbf{Uncertainty estimation} is valuable in many AI scenarios, yet not straightforward to implement in practice. This is further complicated with multiple data sources and users, each injecting generally unknown amounts of noise and uncertainties. In medical applications it is of critical importance, to provide measures of confidence and sensitivity, and for AI researchers through end-users. In anomaly detection, various uncertainty measures can help calibrate the false-positive versus false-negative rates, which can be very domain specific.
    \item \textbf{Costs of edge cases} can be significant, sometimes risking expensive machine downtime or medical failures. This is exacerbated in anomaly detection anomalies are by definition rare so they can be difficult to train for, especially for the anomalies that are completely unseen until they arise in the wild.
    \item \textbf{End-user trust} can be difficult to achieve, often preventing the adoption of ML applications, particularly in the healthcare domain and other highly regulated industries.
\end{itemize}

These and additional ML challenges such as data privacy and interpretability can inhibit ML adoption in clinical practice and industrial settings, but can be mitigated with MLTRL processes.
We'll describe how in the context of the Naud \& Lavin\cite{Naud2020ManifoldsFU} example, which began at level 0 with theoretical ML work on manifold geometries, and at level 5 was directed towards specialized human-machine teaming applications utilizing the same ML method under-the-hood.

\begin{itemize}
    \item \textbf{Levels 0-1 --} From open-ended exploration of data-representation properties in various Riemmanian manifold curvatures, we derived from first principles and empirically identified a property with hyperbolic manifolds: when used as a latent space for embedding data without labels, the geometry organizes the data by it's implicit hierarchical structure. Unsupervised computer vision was identified in reviews as a promising direction for proof-of-principle work.

    \item \textbf{Level 2 --} 
    One approach for validating the earlier theoretical developments was to generate synthetic data to isolate very specific features in data we would expect represented in the latent manifold. The results showed promise for anomaly detection -- using the latent representation of data to automatically identify images that are out-of-the-ordinary (anomalous), and also using the manifold to inspect how they are semantically different. Further, starting with an implicitly probabilistic modeling approach implied uncertainty estimation could be a valuable feature downstream. This made the level 2 key decision point clear: proceed with applied ML development. 

    \item \textbf{Levels 3-5 --} 
    Proof-of-concept development and reviews demonstrated promise for several commercial applications relevant to the business, and also highlighted the need for several key features (defined as R\&D and product requirements): interpretability (towards end-user trust), uncertainty quantification (to show confidence scores), and human-in-the-loop (for domain expertise). Without the MLTRL PoC steps and review processes, these features can often be delayed until beta testing or overlooked completely -- for example, the failures of applying IBM Watson in medical applications \cite{2018TowardsTM}. For this technology, the applications to develop towards are anomaly detection in histopathology and manufacturing, specifically inspecting whole-slide images of neural tissue, and detecting defects in metallic surfaces, respectively.
    
    From the systems perspective, we suggest quantifying the uncertainties of components and propagating them through the system, which can improve safety and trust. Probabilistic ML methods, rooted in Bayesian probability theory, provide a principled approach to representing and manipulating uncertainty about models and predictions\cite{Ghahramani2015ProbabilisticML}. For this reason we advocate strongly for probabilistic models and algorithms in AI systems. 
    In this machine vision example, the MLTRL technical requirements specifically called for a probabilistic generative model to readily quantify various types of uncertainties and propagate them forward to the visualization component of the pipeline, and the product requirements called for the downstream confidence and sensitivity measures to be exposed to the end-user.
    Component uncertainties must be assembled in a principled way to yield a meaningful measure of overall system uncertainty, based on which safe decisions can be made\cite{McAllister2017ConcretePF}. See the Methods section for more on uncertainty in AI systems.

    The early checks for data management and governance proved valuable here, as the application areas dealt with highly sensitive data that would significantly influence the design of data pipelines and test suites. In both the neuropathology and manufacturing applications, the data management checks also raised concerns about hidden feedback loops, where users may unintentionally skew the data inputs when using the anomaly detection models in practice, for instance biasing the data towards specific subsets they subjectively need help with. Incorporating domain experts this early in the project lifecycle helped inform verification and validation steps to help be robust to the hidden feedback loops. Not to mention their input guided us towards user-centric metrics for performance, which can often skew from ML metrics in important ways -- for instance, the typical acceptance ratio for false positives versus false negatives doesn't apply to select edge cases, for which our hierarchical anomaly classification scheme was useful \cite{Naud2020ManifoldsFU}.

    From prior reviews and TRL card documentation, we also identified the value of synthetic data generation into application development: anomalies are by definition rare so they are hard to come by in real datasets, especially with evolving environments in deployment settings, so the ability to generate synthetic datasets for anomaly detection can accelerate the level 6-9 pipeline, and help ensure more reliable models in the wild.

    \item \textbf{Level 6 (medical) --}
    The medical inspection application experienced a bifurcation with product work proceeding while additional R\&D was desired to explore improved data processing methods, while engaging with clinicians and medical researchers for feedback. Proceeding through the levels in a non-linear, non-monotonic way is common in MLTRL and encouraged by various switchback mechanisms (detailed in the Methods section).
    These practices -- intentional switchbacks, frequent engagement with domain experts and users -- can help mitigate methodological flaws and underlying biases that are common when applying ML to clinical applications. For instance, recent work by Roberts et al. \cite{Roberts2021COVID} investigated 2,122 studies applying ML to COVID-19 use-cases, finding that none of the models are sufficient for clinical use due to methodological flaws and/or underlying biases. They go on to give many recommendations -- some we've discussed in the context of MLTRL, and more -- which should be reviewed for higher quality medical-ML models and documentation. 

    \item \textbf{Level 6-9 (manufacturing) --} 
    Overall these stages proceeded regularly and efficiently for the defect detection product. MLTRL's embedded switchback from level 9 to 4 proved particularly useful in this lifecycle, both for incorporating feedback from the field and for updating with research progress. On the former, the data distribution shifts from one deployment setting to another significantly affected false-positive versus false-negative calibrations, so this was added as a feature to the CI/CD pipelines. On the latter, the built-in touch points for real-world feedback and data into the continued ML research provided valuable constraints to help guide research, and product managers could readily understand what capabilities could be available for product integration and when (readily communicated with TRL Cards) -- for instance, later adding support for video-based inspection for defects, and tooling for end-users to reason about uncertainty estimates (which helps establish trust).

    \item \textbf{Level 7-9 (medical) --} 
    For productization the ``neuropathology copilot'' was handed off to a partner pharmaceutical company to integrate into their existing software systems. The MLTRL documentation and communication streamlined the technology transfer, which can often by a time-consuming manual process.
    If not pursuing this path, the product would've likely faced many of the medical-ML deployment challenges with model availability and data access; MLTRL cannot overcome the technical challenges of deploying on-premises, but the manifestation of those challenges as performance regressions, data shifts, privacy and ethics concerns, etc. can be mitigated by the system-level checks and strategies MLTRL puts forth.

\end{itemize}

\begin{figure}[t]
\centering
\includegraphics[width=\linewidth]{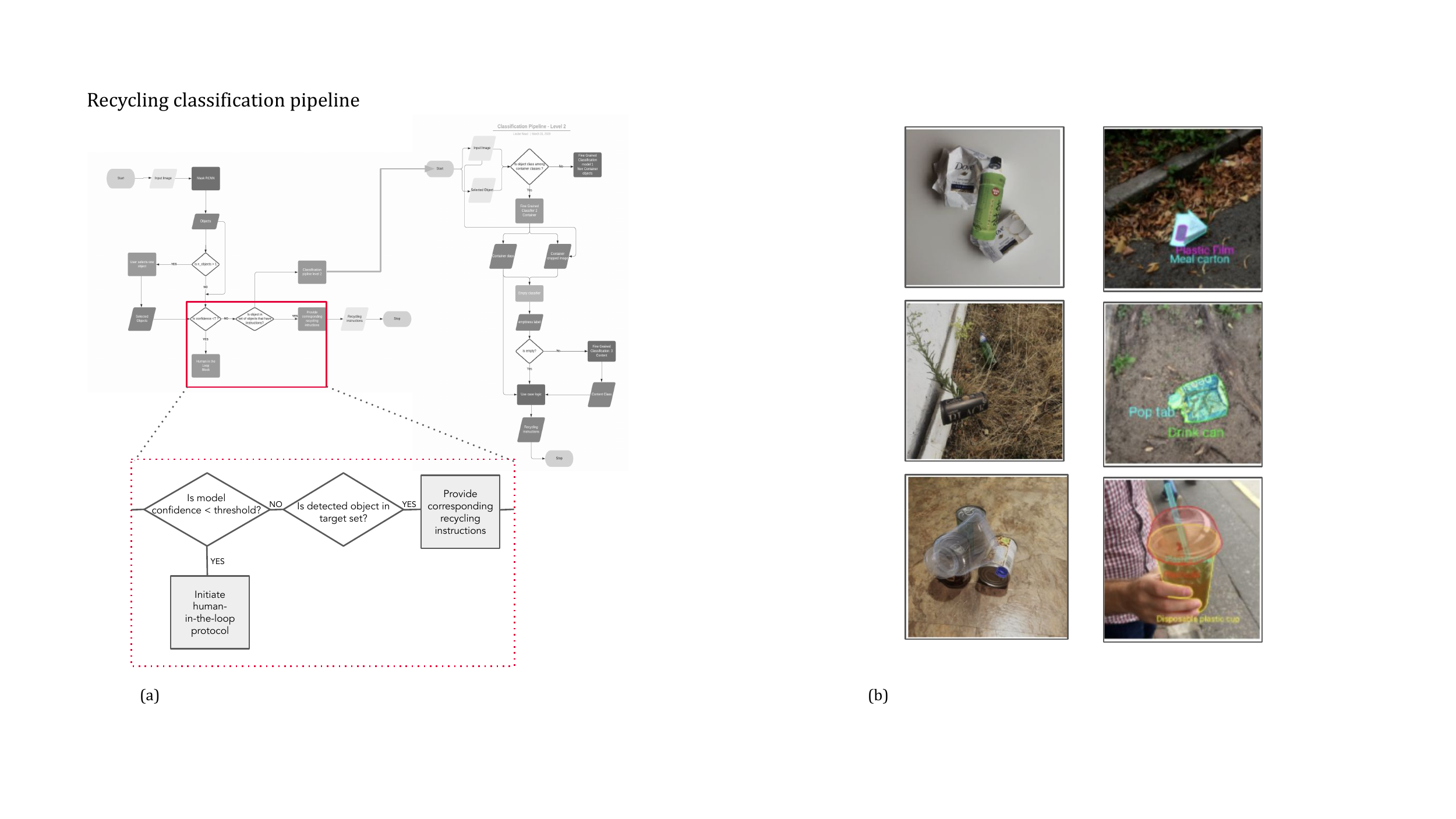}
\caption{
Computer vision pipeline for an automated recycling application (a), which contains multiple ML models, user input, and image data from various sources. Complicated logic such as this can mask ML model performance lags and failures, and also emphasized the need for R\&D-to-product hand off described in MLTRL. Additional emphasis is placed on ML tests that consider the mix of real-world data with user annotations (b, right) and synthetic data generated by Unity AI's Perception tool and structured domain randomization (b, left).
}
\label{fig:unity}
\end{figure}

\subsubsection*{Computer vision with real and synthetic data}
Advancements in physics engines and graphics processing have advanced AI environment and data-generation capabilities, putting increased emphasis on transitioning models across the simulation-to-reality gap~\cite{Tobin2017DomainRF,Juliani2018UnityAG,Hinterstoier2019AnAS}. 
To develop a computer vision application for automated recycling, we leveraged the Unity Perception~\cite{Perception2021} package, a toolkit for generating large-scale datasets for perception-based ML training and validation. We produced synthetic images to complement real-world data sources (Figure \ref{fig:unity}).
This application exemplifies three important challenges in ML product development that MLTRL helps overcome: 

\begin{itemize}
    \item \textbf{Multiple and disparate data sources} are common in deployed ML pipelines yet often ignored in R\&D. For instance, upstream data providers can change formats unexpectedly, or a physical event could cause the customer behavior to change. It is nearly impossible to anticipate and design for all potential problems with real-world data and deployment. This computer vision system implemented pipelines and extended test suites to cover open-source benchmark data, real user data, and synthetic data.  
    
    \item \textbf{Hidden performance degradation} can be challenging to detect and debug in ML systems because gradual changes in performance may not be immediately visible. Common reasons for this challenge are that the ML component may be one step in a series.  Additionally, local/isolated changes to an ML component's performance may not directly affect the observed downstream performance. We can see both issues in the illustrated logic diagram for the automated recycling app (Figure~\ref{fig:unity}). A slight degradation in the initial CV model may not heavily influence the following user input. However, when an uncommon input image appears in the future, the app fails altogether.
    
    \item \textbf{Model usage requirements} can make or break an ML product. For example, the Netflix ``\$1M Prize'' solution was never fully deployed because of significant engineering costs in real-world scenarios~\footnote{\href{https://netflixtechblog.com/netflix-recommendations-beyond-the-5-stars-part-1}{netflixtechblog.com/netflix-recommendations-beyond-the-5-stars-part-1}}.
For example, engineering teams must communicate memory usage, compute power requirements, hardware availability, network privacy, and latency to the ML teams. ML teams often only understand the statistics or ML theory behind a model but not the system requirements or how it scales.
    
\end{itemize}

We next elucidate these challenges and how MLTRL helps overcome them in the context of this project's lifecycle. This project started at level 4, using largely existing ML methods with a target use-case.

\begin{itemize}
    \item \textbf{Level 4 --} For this project, we validated most of the components in other projects. Specifically, the computer vision (CV) model for object recognition and classification was an off-the-shelf model. The synthetic data generation method used Unity Perception, a well-established open-source project. Though this allowed us to skip the earlier levels, many challenges arise when combining ML elements that were independently validated and developed. The MLTRL prototype-caliber code checkpoint ensures that the existing code components are validated and helps avoid poorly defined borders and abstractions between components. ML pipelines often grow out of glue code, and our regimented code checkpoints motivate well-architected software that minimizes these danger spots.

    \item \textbf{Level 5 --} The problematic ``valley of death'', mentioned earlier in the level 5 definitions, is less prevalent in use-cases like this that start at a higher MLTRL level with a specific product deliverable. In this case, the product deliverable was a real-time object recognition and classification of trash for a mobile recycling application. Still, this stage is critical for the requirements and V\&V transition. This stage mitigates failure risks due to the disparate data sources integrated at various steps in this CV system and accounted for the end-user compute constraints for mobile computing. Specifically, the TRL cards from earlier stages surfaced potential issues with imbalanced datasets and the need for specific synthetic images. These considerations are essential for the data readiness and testing V\&V in the productization requirements. Data quality and availability issues often present huge blockers because teams discover them too late in the game. Data-readiness is one class of many example issues teams face without MLTRL, as depicted in Fig. \ref{fig:trl_flows}.
    
    \item \textbf{Level 6 --} 
    We were re-using a well-understood model and deployment pipeline in this use-case, meaning our primary challenge was around data reliability. For the problem of recognizing and classifying trash, building a reliable data source using only real data is almost impossible due to diversity, class imbalance, and annotation challenges. Therefore we chose to develop a synthetic data generator to create training data. At this MLTRL level, we needed to ensure that the synthetic data generator created sufficiently diverse data and exposed the controls needed to alter the data distribution in production. Therefore, we carefully exposed APIs using the Unity Perception package, which allowed us to control lighting, camera parameters, target and non-target object placements and counts, and background textures. Additionally, we ensured that the object labeling matched the real-world annotator instructions and that output data formats matched real-world counterparts. Lastly, we established a set of statistical tests to compare synthetic and real-world data distributions. The MLTRL checks ensured that we understood, and in this case, adequately designed our data sources to meet in-production requirements.  

    \item \textbf{Level 7 --} From the previous level's R\&D TRL cards and observations, we knew relatively early in productization that we would need to assume bias for the real data sources due to class imbalance and imperfect annotations. Therefore we designed tests to monitor these in the deployed application. MLTRL imposes these critical deployment tests well ahead of deployment, where we can easily overlook ML-specific failure modes.  

    \item \textbf{Level 8 --} 
    As we suggested earlier, problems that stem from real-world data are near impossible to anticipate and design for, implying the need for level 8 flight-readiness preparations. Given that we were generating synthetic images (with structured domain randomization) to complement the real data, we created tests for different data distribution shifts at multiple points in the classification pipeline. We also implemented thorough shadow tests ahead of deployment to evaluate how susceptible the ML model(s) to performance regressions caused by data. Additionally, we also implemented these as CI/CD tests over various deployment scenarios (or mobile device computing specifications). Without these fully covered, documented, and automated, it would be impossible to pass level 8 review and deploy the technology.

    \item \textbf{Level 9 --} Post-deployment, the monitoring tests prescribed at Levels 8 and 9 and the three main code quality checkpoints in the MLTRL process help surface hidden performance degradation problems, common with complex pipelines of data flows and various models. The switchbacks depicted in Fig. \ref{fig:trl_flows} are typical in CV use-cases. For instance, miscalibrations in models pre-trained on synthetic data and fine-tuned on newer real data can be common yet difficult to catch. However, the level 7 to 4 switchback is designed precisely for these challenges and product improvements.

\end{itemize}

\subsubsection*{Accelerating scientific discovery with massive particle physics simulators}
Computational models and simulation are key to scientific advances at all scales, from particle physics, to material design and drug discovery, to weather and climate science, and to cosmology\cite{Cranmer2020TheFO}.
Many simulators model the forward evolution of a system (coinciding with the arrow of time), such as the interaction of elementary particles, diffusion of gasses, folding of proteins, or evolution of the universe in the largest scale. 
The task of inference refers to finding initial conditions or global parameters of such systems that can lead to some observed data representing the final outcome of a simulation. In probabilistic programming\cite{Meent2018AnIT}, this inference task is performed by defining prior distributions over any latent quantities of interest, and obtaining posterior distributions over these latent quantities conditioned on observed outcomes (for example, experimental data) using Bayes rule. This process, in effect, corresponds to inverting the simulator such that we go from the outcomes towards the inputs that caused the outcomes. 
In the ``Etalumis'' project\cite{Baydin2019EtalumisBP} (``simulate'' spelled backwards), we are using probabilistic programming methods to invert existing, large-scale simulators via Bayesian inference.
The project is as an interdisciplinary collaboration of specialists in probabilistic machine learning, particle physics, and high-performance computing, all essential elements to achieve the project outcomes. 
Even more, it is a multi-year project spanning multiple countries, companies, university labs, and government research organizations, bringing significant challenges in project management, technology coordination and validation. Aided by MLTRL, there were several key challenges to overcome in this project that are common in scientific-ML projects:

\begin{itemize}
    \item \textbf{Integrating with legacy systems} is common in scientific and industrial use-cases, where ML methods are applied with existing sensor networks, infrastructure, and codebases. In this case, particle physics domain experts at CERN are using the SHERPA simulator\cite{Gleisberg2009EventGW}, a 1 million line codebase developed over the last two decades. Rewriting the simulator for ML use-cases is infeasible due to the codebase size and buried domain knowledge, and new ML experts would need significant onboarding to gain working knowledge of the codebase. It is also common to work with legacy data infrastructure, which can be poorly organized for machine learning (let alone preprocessed and clean) and unlikely to have followed best practices such as dataset versioning.
    \item \textbf{Coupling hardware and software architectures} is non-trivial when deploying ML at scale, as performance constraints are often considered in deployment tests well after model and algorithm development, not to mention the expertise is often split across disjoint teams. This can be exacerbated in scientific-ML when scaling to supercomputing infrastructure, and working with massive datasets that can be in the terabytes and petabytes.
    \item \textbf{Interpretability} is often a desired feature yet difficult to deliver and validate in practice. Particularly in scientific ML applications such as this, mechanisms and tooling for domain experts to interpret predictions and models are key for usability (integrating in workflows and building trust).
\end{itemize}

To this end, we will go through the MLTRL levels one by one, demonstrating how they ensure the above scientific ML challenges are diligently addressed. 

\begin{itemize}
    \item \textbf{Level 0 --} The theoretical developments leading to Etalumis are immense and well discussed in Baydin et al \cite{Baydin2019EtalumisBP}. In particular the ML theory and methods are in a relatively nascent area of ML and mathematics, probabilistic programming. New territory can present more challenges compared to well-traveled research paths, for instance in computer vision with neural networks. It is thus helpful to have a guiding framework when making a new path in ML research, such as MLTRL where early reviews help theoretical ML projects get legs.

    \item \textbf{Level 1-2 --} Running low-level experiments in simple testbeds is generally straightforward when working with probabilistic programming and simulation; in a sense, this easy iteration over experiments is what PPL are designed for. It was additionally helpful in this project to have rich data grounded in physical constraints, allowing us to better isolate model behaviors (rather than data assumptions and noise).
    The MLTRL requirements documentation is particularly useful for the standard PPL experimentation workflow: model, infer, criticize, repeat (or Box's loop) \cite{Blei2014BuildCC}. The evaluation step (i.e. criticizing the model) can be more nuanced than checking summary statistics as in deep learning and similar ML workflows. It is thus a useful practice to write down the criticism methods, metrics, and expected results as verifications for specific research requirements, rather than iterating over Box's loop without \textit{a priori} targets.
    Further, because this research project had a specific target application early in the process (the SHERPA simulator), the project timeline benefited from recognizing simulator-integration constraints upfront as requirements, not to mention data availability concerns, which are often overlooked in early R\&D levels. It was additionally useful to have CERN scientists as domain experts in the reviews at these R\&D levels.
    
    \item \textbf{Level 3 --} Systems development can be challenging with probabilistic programming, again because it is relatively nascent and much of the out-of-the-box tools and infrastructure are not there as in most ML and deep learning. Here in particular there's a novel (unproven) approach for systems integration: a probabilistic programming execution protocol was developed to reroute random number draws in the stochastic simulator codebase (SHERPA) to the probabilistic programming system, thus enabling the system to control stochastic choices in SHERPA and run inference on its execution traces, all while keeping the legacy codebase intact! A more invasive method that modifies SHERPA would not have been acceptable.  
    If it were not for MLTRL forcing systems considerations this early in the Etalumis project lifecycle, this could have been an insurmountable hurdle later when multiple codebases and infrastructures come into play. By the same token, systems planning here helped enable the significant HPC scaling later: the team defined the need for HPC support well ahead of actually running HPC, in order to build the prototype code in a way that would readily map to HPC (in addition to local or cloud CPU and GPU). The data engineering challenges in this system's development nonetheless persist -- that is, data pipelines and APIs that can integrate various sources and infrastructures, and normalize data from various databases -- although MLTRL helps consider these at the an earlier stage that can help inform architecture design.
    
    \item \textbf{Level 4 --} The natural ``embedded switchback'' from Level 4 to 2 (see the Methods section) provided an efficient path toward developing an improved, amortized inference method--i.e., using a computationally expensive deep learning based inference algorithm to train only once, in order to then do fast, repeated inference in the SHERPA model.
    Leveraging cyclic R\&D methods, the Etalumis project could iteratively improve inference methods without stalling the broader system development, ultimately producing the largest scale posterior inference in a Turing-complete probabilistic programming system.
    Achieving this scale through iterative R\&D along the main project lifecycle was additionally enabled by working with with NERSC engineers and their Cori supercomputer to progressively scale smaller R\&D tests to the goal supercomputing deployment scenario. Typical ML workflows that follow simple linear progressions\cite{mlworkflow,Amershi2019SoftwareEF} would not enable ramping up in this fashion, and can actual prevent scaling R\&D to production due to lack of systems engineering processes (like MLTRL) connecting research to deployment.
    
    \item \textbf{Level 5 --} Multi-org international collaborations can be riddled with communication and teamwork issues, in particular at this pivotal stage where teams transition from R\&D to application and product development. First, MLTRL as a lingua franca was key to the team effort bringing Etalumis proof-of-concept into the larger effort of applying it to massive high-energy physics simulators.
    It was also critical at this stage to clearly communicate end-user requirements across the various teams and organizations, which must be defined in MLTRL requirements docs with V\&V measures -- the essential science-user requirements were mainly for model and prediction interpretability, uncertainty estimation, and code usability. If there are concerns over these features, MLTRL switchbacks can help to quickly cycle back and improve modeling choices in a transparent, efficient way -- generally in ML projects, these fundamental issues with usability are caught too late, even after deployment.
    In the probabilistic generative model setting we’ve defined in Etalumis, Bayesian inference gives results that are interpretable because they include exact locations and processes in the model that are associated with each prediction. Working with ML methods that are inherently interpretable, we are well-positioned to deliver interpretable interfaces for the end-users later in the project lifecycle.
    
    \item \textbf{Level 6-9 --} The standard MLTRL protocol apply in these application-to-deployment stages, with several Etalumis-specific highlights. 
    First, given the significant research contributions in both probabilistic programming and scientific-ML, it's important to share the code publicly. The development and deployment of the open-source code repository PPX\footnote{\href{https://github.com/pyprob/ppx}{github.com/pyprob/ppx}} branched into a separate MLTRL path from the Etalumis path for deployment at CERN. It's useful to have systems engineering enable clean separation of requirements, deployments, etc. when there are different development and product lifecycles originating from a common parent project. For example, in this case it was useful to employ MLTRL switchbacks in the open-sourcing process, isolated from the CERN application paths, in order to add support for additional programming languages so PPX can apply to more scientific simulators -- both directions benefited significantly the from the data pipelines considerations brought up levels earlier, where open-sourcing required different data APIs and data transformations to enable broad usability.
    Second, related to the open-source code deliverable and the scientific ML user requirements we noted above, the late stages of MLTRL reviews include higher level stakeholders and specific end-users, yet again enforcing these scientific usability requirements are met. An example result of this in Etalumis is the ability to output human-readable execution traces of the SHERPA runs and inference, enabling never before possible step-by-step interpretability of the black-box simulator.

\end{itemize}

The scientific ML perspective additionally brings to forefront an end-to-end data perspective that is pertinent in essentially all ML use-cases: these systems are only useful to the extent they provide comprehensive data analyses that integrate the data consumed and generated in these workflows, from raw domain data to machine-learned models. These data analyses drive reproducibility, explainability, and experiment data understanding, which are critical requirements in scientific endeavors and ML broadly.

\subsubsection*{Causal inference \& ML in medicine}

Understanding cause and effect relationships is crucial for accurate and actionable decision-making in many settings, from healthcare and epidemiology, to economics and government policy development. Unfortunately, standard machine learning algorithms can only find patterns and correlation in data, and as correlation is not causation, their predictions cannot be confidently used for understanding cause and effect. Indeed, relying on correlations extracted from observational data to guide decision-making can lead to embarrassing, costly, and even dangerous mistakes, such as concluding that asthma reduces pneumonia mortality risk~\cite{Ambrosino1995TheUO}, and that smoking reduces risk of developing severe COVID-19~\cite{griffith2020collider}.
Fortunately, there has been much recent development in a field known as causal inference that can quantitatively make sense of cause and effect from purely observational data\cite{Pearl2018TheoreticalIT}. The ability of causal inference algorithms to quantify causal impact rests on a number of important checks and assumptions--beyond those employed in standard machine learning or purely statistical methodology--that must be carefully deliberated over during their development and training. These specific checks and assumptions are as follows:
\begin{itemize}
    \item \textbf{Specifying cause-and-effect relationships between relevant variables--}
 One of the most important assumptions underlying causal inference is the structure of the causal relations between quantities of interest. The gold standard for determining causal relations is to perform a randomised controlled trial, but in most cases these cannot be employed due to ethical concerns, technological infeasibility, or prohibitive cost. In these situations, domain experts have to be consulted to determine the causal relationships. It is important in these situations to carefully address the manner in which such domain knowledge was extracted from experts, the number and diversity of experts involved, the amount of consensus between experts, and so on. The need for careful documentation of this knowledge and its periodic review is made clear in the MLTRL framework, as we shall see below.
\item \textbf{Identifiability--} Another vital component of building causal models is whether the causal question of interest is \textit{identifiable} from the causal structure specified for the model together with observational (and sometimes experimental) data. 
\item \textbf{Adjusting for and monitoring confounding bias--} An important aspect of causal model performance, not present in standard machine learning algorithms, is \textit{confounding bias} adjustment. The standard approach is to employ propensity score matching to remove such bias. However, the quality of bias adjustment achieved in any specific instance with such propensity-based matching methods needs to be checked and documented, with alternate bias adjusting procedure required if appropriate levels of bias adjustment are not achieved\cite{Nguyen2017DoubleadjustmentIP}.
\item \textbf{Sensitivity analysis--} As causal estimates are based on generally untestable assumptions, such as observing all relevant confounders, it is vital to determine how sensitive the resulting predictions are to potential violations of these assumptions. 
\item \textbf{Consistency--} It is crucial to understand if the learned causal estimate  provably converges to the true causal effect in the limit of infinite sample size. However, causal models cannot be validated by standard held-out tests, but rather require randomization or special data collection strategies to evaluate their predictions~\cite{Eckles2017BiasAH,Xu2020SplitTreatmentAT}.
\end{itemize}
The MLTRL framework makes transparent the need to carefully document and defend these assumptions, thus ensuring the safe and robust creation, deployment, and maintenance of causal models. We elucidate this with recent work by Richens et al.\cite{Richens2020ImprovingTA}, developing a causal approach to computer-assisted diagnosis which outperforms previous purely machine learning based methods. To this end, we will go through the MLTRL levels one by one, demonstrating how they ensure the above specific checks and assumptions are naturally accounted for. This should provide a blueprint for how to employ the MLTRL levels in other causal inference applications. 
\begin{itemize}
\item \textbf{Level 0 --} When initially faced with a causal inference task, the first step is always to understand the causal relationships between relevant variables. For instance, in Richens et al. \cite{Richens2020ImprovingTA}, the first step toward building the diagnostic model was specifying the causal relationships between the diverse set risk factors, diseases, and symptoms included in the model. To learn these relations, doctors and healthcare professionals were consulted to employ their expansive medical domain knowledge which was robustly evaluated by additional independent groups of healthcare professionals. The MLTRL framework ensured this issue is dealt with and documented correctly, as such knowledge is required to progress from Level 0; failure to do this has plagued similar healthcare AI projects~\cite{Paleyes2020ChallengesID}. 

The next step of any causal analysis is to understand whether the causal question of interest is uniquely identifiable from the causal structure specified for the model together with observational and experimental data. In this medical diagnosis example, identification was crucial to establish, as the causal question of interest, ``would the observed symptoms not be present had a specific disease been cured?'', was highly non-trivial. Again, MLTRL ensures this vital aspect of model building is carefully considered, as a mathematical proof of identifiability would be required to graduate from Level 0.

With both the causal structure and identifiability result in hand, one can progress to Level 1.

\item \textbf{Level 1 --} At this level, the goal is to take the estimand for the identified causal question of interest and devise a way to estimate it from data. To do this one will need efficient ways to adjust for confounfing bias. The standard approach is to employ propensity score-based methods to remove such bias when the target decision is binary, and use multi-stage ML models adhering to the assumed causal structure\cite{Chernozhukov2018DoubleDebiasedML} for continuous target decisions (and high-dimensional data in general). However, the quality of bias adjustment achieved in any specific instance with propensity-based matching methods needs to be checked and documented, with alternate bias adjusting procedure required if appropriate levels of bias adjustment are not achieved\cite{Nguyen2017DoubleadjustmentIP}. As above, MLTRL ensures transparency and adherence to this important aspect of causal model development, as without it a project cannot graduate from Level 1. Even more, MLTRL ensures tests for confounding bias are developed early-on and maintained throughout later stages to deployment. Still, in many cases, it is not possible to completely remove confounding in the observed data. TRL Cards offer a transparent way to declare specific limitations of a causal ML method. 

\item \textbf{Level 2 --} PoC-level tests for causal models must go beyond that of typical ML models. As discussed above, to ensure the estimated causal effects are robust to the assumptions required for their derivation, sensitivity to these assumptions must be analysed. Such sensitivity analysis is often limited to R\&D experiments or a post-hoc feature of ML products. MLTRL on the other hand requires this throughout the lifecycle as components of ML test suites and gated reviews. In the case of causal ML, best practice is to employ sensitivity analysis for this robustness check\cite{veitch2020sense}. MLTRL ensures this check is highlighted and adhered to, and no model will end up graduating Level 2--let alone being deployed--unless it is passed. 

\item \textbf{Level 3 --} Coding best practices, as in general ML applications.

\item \textbf{Level 4-5 --} There are additional tests to consider when taking causal models from research to production, in particular at Level 4--proof of concept demonstration in a real scenario. \textit{Consistency}, for example, is an important property of causal methods that informs us whether the method provably converges to the true causal graph in the limit of infinite sample size. Quantifying consistency in the test suite is critical when datasets change from controlled laboratory settings to open-world, and when the application scales. And PoC validation steps are more efficient with MLTRL because the process facilitates early specification of the evaluation metric for a causal model in Level 2. Causal models cannot be validated by standard held-out tests, but rather require randomization or special data collection strategies to evaluate their predictions\cite{Eckles2017BiasAH,Xu2020SplitTreatmentAT}. Any difficulty in evaluating the model's predictions will be caught early and remedied.

\item \textbf{Level 6-9 --} With the the causal ML components of this technology developed reliably in the previous levels, the rest of the levels developing this technology focused on general medical-ML deployment challenges. For the most part, data governance, privacy, and management that was detailed earlier in the neuropathology MLTRL use-case, as well as on-premises deployment.

\end{itemize}

\subsubsection*{AI for open-source space sciences}

The CAMS (Cameras for Allsky Meteor Surveillance) project~\cite{JENNISKENS201140}, established in 2010 by NASA, uses hundreds of off-the-shelf CCTV cameras to capture the meteor activity in the night sky. Initially, resident scientists would retrieve hard-disks containing video data captured each night and perform manual triangulation of tracks or streaks of light in the night sky, and compute a meteor's trajectory, orbit, and lightcurve. Each solution was manually classified as a meteor or not (i.e., planes, birds, clouds, etc). In 2017, a project run by the Frontier Development Lab\footnote{The NASA Frontier Development Lab and partners open-source the code and data via the SpaceML platform: \href{http://spaceml.org/}{spaceml.org}}\cite{Ganju2020LearningsFF}, the AI accelerator for NASA and ESA, aimed to automate the data processing pipeline and replicate the scientists thought process to build an ML model that identifies meteors in the CAMS project~\cite{Zoghbi2017SearchingFL, JENNISKENS201821}. The data automation led to orders of magnitude improvements in operational efficiency of the system, and allowed new contributors and amateur astronomers to start contributing to meteor sightings. Additionally, a novel web tool allowed anybody anywhere to view the meteors detected in the previous night. The CAMS camera system has had six-fold global expansion of the data capture network, discovered ten new meteor showers, contributed towards instrumental evidence of previously predicted comets, and helped calculate parent bodies of various meteor showers. CAMS utilized the MLTRL framework to progress as described:

\begin{itemize}
    \item \textbf{Level 1 --} Understanding the domain and data is a prerequisite for any ML development. Extensive data exploration elucidated visual differences between objects in the night sky such as meteors, satellites, clouds, tail lights of planes, light from the eyes of cats peering into cameras, trees, and other tall objects visible in the moonlight. This step helped (1) understand visual properties of meteors that later defined the ML model architecture, and (2) mitigate impact of data imbalance by proactively developing domain-oriented strategies. 
    The results are well-documented on a datasheet associated with the TRL card, and discussed at the stage review. This MLTRL documentation forced us to consider data sharing and other privacy concerns at this early conceptualization stage, which is certainly relevant considering CAMS is for open-source and gathering data from myriad sources.
    \item \textbf{Level 2-3 --} The agile and non-monotonic (or non-linear) development prescribed by MLTRL allowed the team to first develop an approximate end-to-end pipeline that offered a path to ML model deployment and quick turnaround time to incorporate feedback from the regular gated reviews. Then, with relatively quicker experimentation, the team could improve on the quality of not just the ML model, but also scale up the systems development simultaneously in a non-monotonic development cycle.
    \item \textbf{Level 4 --} With the initial pipeline in place, scalable training of baselines and initial models on real challenging datasets ensued. 
    Throughout the levels, the MLTRL gated reviews were essential for making efficient progress while ensuring robustness and functionality that meets stakeholder needs.
    At this stage we highlight specific advantages of the MLTRL review processes that had instrumental effect on the project success:
    With the required panel of mixed ML researchers and engineers, domain scientists, and product managers, the stage 4 reviews stressed the significance of numerical improvements and comparison to existing baselines, and helped identify and overcome issues with data imbalance. The team likely would have overlooked these approaches without the review from peers in diverse roles and teams.
    In general, the evolving panel of reviewers at different stages of the project was essential for covering a variety of verification and validation measures -- from helping mitigate data challenges, to open-source code quality. 
    \item \textbf{Level 5 --} To complete this R\&D-to-productization level, a novel web tool called the NASA CAMS Meteor Shower Portal\footnote{\href{https://meteorshowers.seti.org/}{meteorshowers.seti.org}} was created that allowed users to view meteor shower activity from the previous night and verify meteor predictions generated by the ML model.
    This app development was valuable for A/B testing, validating detected meteors and classified new meteor showers with human-AI interaction, and demonstrating real-world utility to stakeholders in review. ML processes without MLTRL miss out on these valuable development by overlooking the need for such a demo tool.
    
    \item \textbf{Level 6 --} Application development was naturally driven by end-user feedback from the web app in level 5 -- without MLTRL it's unlikely the team would be able to work with early productization feedback.
    With almost real time feedback coming in daily, newer methods for improving robustness of meteor identification led to researching and developing a unique augmentation technique, resulting in the state of the art performance of the ML model.
    Further application development led to incorporating features that were in demand by users of the NASA CAMS Meteor Shower Portal: include celestial reference points through constellations, add ability to zoom in/out and (un)cluster showers, and provide tooling for scientific communication. 
    The coordination of these features into product-caliber codebase resulted in the release of the NASA CAMS Meteor Shower Portal 2.0 that was built by a team of citizen scientists -- again we found the specific checkpoints in the MLTRL review were crucial for achieving these goals.

    \item \textbf{Level 7 --} 
    Integration was particularly challenging in two ways. First, integrating the ML and data engineering deliverables with the existing infrastructure and tools of the larger CAMS system, which had started development years earlier with other teams in partner organizations, required quantifiable progress for verifying the tech-readiness of ML models and modules. The use of technology readiness levels provided a clear and consistent metric for the maturity of the ML and data technologies, making for clear communication and efficient project integration. Without MLTRL it is difficult to have a conversation, let alone make progress, towards integrating AI/ML and data subsystems and components.
    Second, integrating open-source contributions into the main ML subsystem was a significant challenge alleviated with diligent verification and validation measures from MLTRL, as well as quantifying robustness with ML testing suites (using scoring measures like that of the ML Testing Rubric\cite{Breck2017TheMT}, and devising a checklist based on metamorphic testing\cite{Xie2011TestingAV}).
    
    \item \textbf{Level 8 --} CAMS, like many datasets in practice, consisted of a smaller labeled subset and a much larger unlabeled set. In an attempt to additionally increase robustness of the ML subsystem ahead of ``flight readiness'', we looked to active learning \cite{Cohn1994ActiveLW, Gal2017DeepBA} techniques to leverage the unlabeled data. 
    Models using an initial version of this approach, where results of the active learning provided ``weak'' labels, resulted in consumption of the entire decade long unlabelled data collected by CAMS and slightly higher scores on deployment tests. Active learning showed to be a promising feature and was \textit{switched back} to level 7 for further development towards the next deployment version, so as not to delay the rest of the project.
    
    \item \textbf{Level 9 --} The ML components in CAMS require continual monitoring for model and data drifts, such as changes in weather, smoke, and cloud patterns that affect the view of the night sky. The data drifts may also be specific to locations, such as fireflies and bugs in CAMS Australia and New Zealand stations which appear as false positives. The ML pipeline is largely automated with CI/CD, runs regular regression tests, and production of benchmarks. Manual intervention can be triggered when needed, such as sending low confidence meteors for verification to scientists in the CAMS project. The team also regularly releases the code, models, and web tools on the open-source space sciences and exploration ML toolbox, SpaceML\footnote{\href{http://spaceml.org/}{spaceml.org}}. Through the SpaceML community and partner organizations, CAMS continually improves with feature requests, debugging, and improving data practices, while tracking progress with standard software release cycles and MLTRL documentation.
\end{itemize}

\subsection*{BEYOND SOFTWARE ENGINEERING}

Software engineering (SWE) practices vary significantly across domains and industries. Some domains, such as medical applications, aerospace, or autonomous vehicles rely on a highly rigorous development process which is required by regulations. Other domains, for example advertising and e-commerce are not regulated and can employ a lenient approach to development. ML development should at minimum inherit the acceptable software engineering practices of the domain. There are, however, several key areas where ML development stands out from SWE, adding its own unique challenges which even most rigorous SWE practices are not able to overcome.

For instance, the behavior of ML systems is learned from data, not specified directly in code. The data requirements around ML (i.e., data discovery, management, and monitoring) adds significant complexity not seen in other types of SWE. There are many benefits to using a \textit{data-oriented architecture (DOA)} \cite{Paleyes2020ChallengesID} with the data-first workflows and management practices prescribed in MLTRL. DOA aims to make the data flowing between elements of business logic more explicit and accessible with a streaming-based architecture rather than the micro-service architectures that are standard in software systems. One specific benefit of DOA is making data available and traceable by design, which helps significantly in the ML logging challenges and data governance needs we discussed in Levels 7-9. Moreover, MLTRL highlights data-related requirements along every step to ensure that the development process considers data readiness and availability. 

Not to mention an array of ML-specific failure modes; for example, models that become miscalibrated due to subtle data distributional shifts in the deployment setting, resulting in models that are more confident in predictions than they should be. MLTRL helps define ML-specific testing considerations (levels 5 and 7) to help surface these failure-modes early. 
ML opens up new threat vectors across the whole deployment workflow that otherwise aren't risks in software systems: for example, a poisoning attack to contaminate the training phase of ML systems, or membership inference to see if a given data record was part of the model's training. MLTRL consider these threat vectors and suggests relevant risk-identification during prototyping and productization phases. 
More generally, ML codebases have all the problems for regular code, plus ML-specific issues at the system level, mainly as a consequence of added complexity and dynamism. The resulting entanglement, for instance, implies that the SWE practice of making isolated changes is often not feasible -- Scully et al.\cite{Sculley2015HiddenTD} refer to this as the  ``changing anything changes everything'' principle. Given this consideration, typical SWE change-management is insufficient. Furthermore, ML systems almost necessarily increase the technical debt; package-level refactoring is generally sufficient for removing technical debt in software systems, but this is not the case in ML systems.

These factors and others suggest that inherited software engineering and management practices of a given domain are insufficient for the successful development of robust and reliable ML systems. But it is not trading off one for the other: MLTRL can be used in synergy with the existing, industry-standard software engineering practices such as agile \cite{Abrahamsson2017AgileSD} and waterfall \cite{Kuhrmann2017HybridSA} to handle unique challenges of ML development.
Because ML applications are a category of software, all best practices of building and operating software should be extended when possible to the ML application. Practices like version control, comprehensive testing, continuous integration and continuous deployment are all applicable to ML development. MLTRL provides a framework that helps extend SWE building and operating practices that are acceptable in a given domain to tackle the unique challenges of ML development.

\subsection*{RELATED WORKS}

A recent case study from Microsoft Research~\cite{Amershi2019SoftwareEF} similarly identifies a few themes describing how ML is not equal to software engineering, and recommends a linear ML workflow with steps for data preparation through modeling and deploying.
They define an effective workflow for isolated development of an ML model, but this approach does not ensure the technology is actually improving in quality and robustness. Their process should be repeated at progressive stages of development in the broader ML and data technology lifecycle. If applied in the MLTRL framework, the specific ingredients of the ML model workflow -- that is, people, software, tests, objectives, etc. -- evolve over time and subsequent stages as the technologies mature. 

There exist many recommended workflows for specific ML methods and areas of pipelines. For instance, a more iterative process for Bayesian ML \cite{Gelman2020Bayesian} and even more specifically for probabilistic programming~\cite{Blei2014BuildCC}, a data mining process defined in 2000 that remains widely used~\cite{Chapman2000CRISPDM1S}, others for describing data iterations \cite{Hohman2020UnderstandingAV}, and human-computer interaction cycles \cite{Amershi2014PowerTT}.
In these recommended workflows and others, there's an important distinction between their cycles and ``switchback'' mechanisms in MLTRL. Their cycles suggest to generically iterate over a data-modeling-evaluation-deployment process. Switchbacks, on the other hand, are specific, purpose-driven workflows for dialing part(s) of a project to an earlier stage -- this doesn't simply mean go back and train the model on more data, but rather switching back regresses the technology's maturity level (e.g. from level 5 to level 3) such that it must again fulfill the level-by-level requirements, evaluations and reviews. See the Methods section for more details on MLTRL switchbacks.
In general, iteration is an important part of data, ML, and software processes. MLTRL is unique from the other recommended processes in many ways, and perhaps most importantly because it considers data flows and ML models in the context of larger systems. These isolated processes (that are specific to e.g. modeling in prototype development or data wrangling in application development) are synergistic with MLTRL because they can be used within each level of the larger lifecycle or framework. For example, the Bayesian modeling processes~\cite{Gelman2020Bayesian,Blei2014BuildCC} we mentioned above are really useful to guide developers of probabilistic ML approaches. But there are important distinctions between executing these modeling steps and cycles in a well-defined prototyping environment with curated data and minimal responsibilities, versus a production environment riddled with sparse and noisy data, that interacts with the physical world in non-obvious ways, and can carry expensive (even hidden) consequences. MLTRL provides the necessary, holistic context and structure to use these and other development processes reliably and responsibly.

Also related to our work, Google teams have proposed ML testing recommendations \cite{Breck2017TheMT} and validating the data fed into ML systems \cite{Breck2019DataVF}.
For NLP applications, typical ML testing practices struggle to translate to real-world settings, often overestimating performance capabilities. An effective way to address this is devising a checklist of linguistic capabilities and test types, as in Ribeiro et al.\cite{Ribeiro2020BeyondAB}--interestingly their test suite was inspired by metamorphic testing, which we suggested earlier in Level 7 for testing systems AI integrations.
A survey by Paleyes et al.~\cite{Paleyes2020ChallengesID} go over numerous case studies to discuss challenges in ML deployment. They similarly pay special attention to the need for ethical considerations, end-user trust, and extra security in ML deployments. On the latter point, Kumar et al.~\cite{Kumar2019FailureMI} provide a table thoroughly breaking down new threat vectors across the whole ML deployment workflow (some of which we mentioned above).
These works, notably the ML security measures and the quantification of an ML test suite in a principled way -- i.e., that does not use misguided heuristics such as code coverage -- are valuable to include in any ML workflow including MLTRL, and are synergistic with the framework we've described in this paper.
These analyses provide useful insights, but they do not provide a holistic, regimented process for the full ML lifecycle from R\&D through deployment.
An end-to-end approach is suggested by Raji et al.\cite{Raji2020ClosingTA}, but only for the specific task of auditing algorithms; components of AI auditing are mentioned in Level 7, and covered throughout in the review processes.

Sculley et al.\cite{Sculley2015HiddenTD} go into more ML debt topics such as undeclared consumers and data dependencies, and go on to recommend an ML Testing Rubric as a production checklist \cite{Breck2017TheMT}. 
For example, testing models by a canary process before serving them into production. This, along with similar shadow testing we mentioned earlier, are common in autonomous ML systems, notably robotics and autonomous vehicles.
They explicitly call out tests in four main areas (ML infrastructure, model development, features and data, and monitoring of running ML systems), some of which we discussed earlier. For example, tests that the training and serving features compute the same values; a model may train on logged processes or user input, but is then served on a live feed with different inputs. 
In addition to the Google ML Testing Rubric, we advocate \textit{metamorphic testing}: 
a SWE methodology for testing a specific set of relations between the outputs of multiple inputs.
True to the checklists in the Google ML Testing Rubric and in MLTRL, metamorphic testing for ML can have a codified list of metamorphic relations\cite{Xie2011TestingAV}.

In domains such as healthcare there have been the introduction of similar checklists for data readiness -- for example, to ensure regulatory‐grade real-world-evidence (RWE) data quality \cite{Miksad2018HarnessingTP} -- yet these are nascent and not yet widely accepted.
Applying AI in healthcare has led to developing guidance for regulatory protocol, which is still a work in progress. Larson et al.\cite{Larson2020RegulatoryFF} provide a comprehensive analysis for medical imaging and AI, arriving at several regulatory framework recommendations that mirror what we outline as important measures in MLTRL: e.g., detailed task elements such as pitfalls and limitations (surfaced on TRL Cards), clear definition of an algorithm relative to the downstream task, defining the algorithm ``capability'' (Level 5), real-world monitoring, and more.

D'amour et al.\cite{DAmour2020UnderspecificationPC} dive into the problem we noted earlier about model miscalibration. 
They point to the trend in machine learning to develop models relatively isolated from the downstream use and larger system, resulting in underspecification that handicaps practical ML pipelines. 
This is largely problematic in deep learning pipelines, but we've also noted this risk in the case of causal inference applications.
Suggested remedies include \textit{stress tests}--empirical evaluations that probe the model's inductive biases on practically relevant dimensions--and in general the methods we define in Level 7. 

\subsection*{LIMITATIONS, RESPONSIBILITIES, and ETHICS}

MLTRL has been developed, deployed, iterated, and validated in myriad environments, as demonstrated by the previous examples and many others. Nonetheless we strongly suggest that MLTRL not be viewed as a cure-all for machine learning systems engineering. Rather, MLTRL provides mechanisms to better enable ML practitioners, teams, and stakeholders to be diligent and responsible with these technologies and data. That is, one cannot implement MLTRL in an organization and turn a blind eye to the many data, ML, and integration challenges we've discussed here. MLTRL is analogous to a pilot's checklist, not autopilot.

MLTRL is intended to be complimentary to existing software development methodologies, not replace or alter them. Specifically, whether the team uses agile or waterfall methods, MLTRL can be adopted to help define and structure phases of the project, as well as the success criteria of each stage. In context of the software development process, the purpose of MLTRL is to help the team minimize the technical dept and risk associated with the delivery of an ML application by helping the development team ask necessary questions.

We discussed many data challenges and approaches in the context of MLTRL, and should highlight again the importance of data considerations in any ML initiative. The data availability and quality can severely limit the ability to develop and deploy ML, whether MLTRL is used or not. 
It is again the responsibility of the ML practitioners, teams, and stakeholders to gather, use, and distribute data in safe, legal, ethical ways. MLTRL helps do so with rigor and transparency, but again is not a solution for data bias. We recommend these recent works on data bias in ML: \cite{Mehrabi2019ASO,Ntoutsi2020BiasID,Jo2020LessonsFA,Wiens2020DiagnosingBI,Challen2019ArtificialIB}.
Further, AI/ML ethics is a continuously evolving, multidisciplinary space -- see \cite{Leslie2019UnderstandingAI}. MLTRL aims to prioritize ethics considerations at each level of the framework, and would do well to also evolve over time with the broader AI/ML ethics developments.

\subsection*{CONCLUSION}

We've described \textit{Machine Learning Technology Readiness Levels (MLTRL)}, an industry-hardened systems engineering framework for robust, reliable, and responsible machine learning. 
MLTRL is derived from the processes and testing standards of spacecraft development, yet lean and efficient for ML, data, and software workflows.
Examples from several organizations across industries demonstrate the efficacy of MLTRL for AI and ML technologies, from research and development through productization and deployment, in important domains such as healthcare and physics, with emphasis on data readiness amongst other critical challenges.
Our aim is MLTRL works in synergy with recent approaches in the community focused on diligent data-readiness, privacy and security, and ethics.
Even more, MLTRL establishes a much-needed lingua franca for the AI ecosystem, and broadly for AI in the worlds of science, engineering, and business.
Our hope is that our systems framework is adopted broadly in AI and ML organizations, and that ``technology readiness levels'' becomes common nomenclature across AI stakeholders -- from researchers and engineers to sales-people and executive decision-makers.

\section*{Methods}\label{sec_methods}

\subsection*{Gated reviews}
At the end of each stage is a dedicated review period: (1) Present the technical developments along with the requirements and their corresponding verification measures and validation steps, (2) make key decisions on path(s) forward (or backward) and timing, and (3) debrief the process\footnote{MLTRL should include regular debriefs and meta-evaluations such that process improvements can be made in a data-driven, efficient way (rather than an annual meta-review). MLTRL is a high-level framework that each organization should operationalize in a way that suits their specific capabilities and resources.}.
As in the gated reviews defined by TRL used by NASA, DARPA, et al., MLTRL stipulates specific criteria for review at each level, as well as calling out specific \textit{key decision points} (noted in the level descriptions above).
The designated reviewers will ``graduate'' the technology to the next level, or provide a list of specific tasks that are still needed (ideally with quantitative remarks).
After graduation at each level, the working group does a brief post-mortem; we find that a quick day or two pays dividends in cutting away technical debt and improving team processes.
Regular gated reviews are essential for making efficient progress while ensuring robustness and functionality that meets stakeholder needs.
There are several important mechanisms in MLTRL reviews that are specifically useful with AI and ML technologies:
First, the review panels evolve over a project lifecycle, as noted below.
Second, MLTRL prescribes that each review runs through an AI ethics checklist defined by the organization; it is important to repeat this at each review, as the review panel and stakeholders evolve considerably over a project lifecycle.
As previously described in the levels definitions, including ethics reviews as an integral part of early system development is essential for informing model specifications and avoiding unintended biases or harm\cite{Obermeyer2019DissectingRB} after deployment.

\subsection*{TRL ``Cards''}

In Figure \ref{fig:trl_card} we succinctly showcase a key deliverable: \textit{TRL Cards}.
The model cards proposed by Google~\cite{Mitchell2019ModelCF} are a useful development for external user-readiness with ML.
On the other hand, our TRL Cards aim to be more information-dense, like datasheets for medical devices and engineering tools -- see the open-source TRL Card repo for examples and templates (to be released at \href{https://github.com/alan-turing-institute}{github.com/alan-turing-institute}). 
These serve as ``report cards'' that grow and improve upon graduating levels, and provide a means of inter-team and cross-functional communication.
The content of a TRL Card is roughly in two categories: project info, and implicit knowledge. The former clearly states info such as project owners and reviewers, development status, and semantic versioning--not just for code, also for models and data. In the latter category are specific insights that are typically siloed in the ML development team but should be communicated to other stakeholders: modeling assumptions, dataset biases, corner cases, etc.
With the spread of AI and ML in critical application areas, we are seeing domain expert consortiums defining AI reporting guidelines -- e.g., Rivera et al.\cite{Rivera2020GuidelinesFC} calling for clinical trials reports for interventions involving AI -- which will greatly benefit from the use of our TRL reporting cards.
We stress that these TRL Cards are key for the progression of projects, rather than documentation afterthoughts.
The TRL Cards thus promote transparency and trust, within teams and across organizations.
TRL Card templates will be open-sourced upon publication of this work, including methods for coordinating use with other reporting tools such as ``Datasheets for Datasets''~\cite{Gebru2018DatasheetsFD}.

\subsection*{Risk mitigation}

Identifying and addressing risks in a software project is not a new practice. However, akin to the MLTRL roots in spacecraft engineering, risk is a ``first-class citizen'' here. In the definition of technical and product requirements, each entry has a calculation of the form $risk = p(failure) \times value$, where the value of a component is an integer $1-10$. Being diligent about quantifying risks across the technical requirements is a useful mechanism for flagging ML-related vulnerabilities that can sometimes be hidden by layers of other software. 
MLTRL also specifies that risk quantification and testing strategies are required for sim-to-real development. That is, there is nearly always a non-trivial gap in transferring a model or algorithm from a simulation testbed to the real world. Requiring explicit sim-to-real testing steps in the workflow helps mitigate unforeseen (and often hazardous) failures.
Additionally, comprehensive ML test coverage that we mention throughout this paper is a critical strategy for mitigating risks \textit{and} uncertainties: ML-based system behavior is not easily specified in advance, but rather depends on dynamic qualities of the data and on various model configuration choices\cite{Breck2017TheMT}.

\subsection*{Non-monotonic, non-linear paths} 

We observe many projects benefit from cyclic paths, dialing components of a technology back to a lower level. 
Our framework not only encourages cycles, we make them explicit with ``switchback mechanisms'' to regress the maturity of specific components in an AI system:
\begin{enumerate}
  \item \textit{Discovery switchbacks} occur as a natural mechanism -- new technical gaps are discovered through systems integration, sparking later rounds of component development\cite{Szajnfarber2014ManagingII}. These are most common in the R\&D levels, for example moving a component of a proof-of-concept technology (at Level 4) back to proof-of-principle development (Level 2).
  \item \textit{Review switchbacks} result from gated reviews, where specific components or larger subsystems may be dialed back to earlier levels.
  This switchback is one of the ``key decision points'' in the MLTRL project lifecycle (as noted in the Levels definitions), and is often a decision driven by business-needs and timing rather than technical concerns (for instance when mission priorities and funds shift).
  This mechanism is common from Level 6/7 to 4, which stresses the importance of this R\&D to product transition phase (see Figure \ref{fig:trl_flows} (left)).
  \item \textit{Embedded switchbacks} are predefined in the MLTRL process. For example, a predefined path from 4 to 2, and from 9 to 4. In complex systems, particularly with AI technologies, these built-in loops help mitigate technical debt and overcome other inefficiencies such as noncomprehensive V\&V steps.
\end{enumerate}
Without these built-in mechanisms for cyclic development paths, it can be difficult and inefficient to build systems of modules and components at varying degrees of maturity. 
Contrary to traditional thought that switchback events should be suppressed and minimized, in fact they represent a natural and necessary part of the complex technology development process -- efforts to eliminate them may stifle important innovations without necessarily improving efficiency.
This is a fault of the standard monotonic approaches in AI/ML projects, stage-gate processes, and even the traditional TRL framework.

It is also important to note that most projects do not start at Level 0; very few ML companies engage in this low-level theoretical research. For example, a team looking to use an off-the-shelf object recognition model could start that technology at Level 3, and proceed with thorough V\&V for their specific datasets and use-cases. However, no technology can skip levels after the MLTRL process has been initiated.
The industry default (that is, without implementing MLTRL) is to ignorantly take pretrained models, run fine tuning on their specific data, and jump to deployment, effectively skipping Levels 5 to 7.
Additionally, we find it is advantageous to incorporate components from other high-TRL ranking projects while starting new projects; MLTRL makes the verification and validation (V\&V) steps straightforward for integrating previously developed ML components.

\subsection*{Evolving people, objectives, and measures}
As suggested earlier, much of the practical value of MLTRL comes at the transition between levels. More precisely, MLTRL manages these oft neglected transitions explicitly as evolving teams, objectives, and deliverables.
For instance, the team (or working group) at Level 3 is mostly AI Research Engineers, but at Level 6 is mixed Applied AI/SW Engineers mixed with product managers and designers.
Similarly, the review panels evolve from level to level, to match the changing technology development objectives.
What the reviewers reference similarly evolves: notice in the level definitions that technical requirements and V\&V guide early stages, but at and after Level 6 the product requirements and V\&V takeover -- naturally, the risk quantification and mitigation strategies evolve in parallel.
Regarding the deliverables, notably TRL Cards and risk matrices\cite{Duijm2015RecommendationsOT} (to rank and prioritize various science, technical, and project risks), the information develops and evolves over time as the technology matures.

\subsection*{Quantifiable progress}
By defining technology maturity in a quantitative way, MLTRL enables teams to accurately and consistently define their ML progress metrics. Notably industry-standard ``objectives and key results'' (OKRs) and ``key performance indicators'' (KPIs)~\cite{Zhou2018ComparativeSO} can be defined as achieving certain readiness levels in a given period of time; this is a preferable metric in essentially all ML systems which consist of much more than a single performance score to measure progress. 
Even more, meta-review of MLTRL progress over multiple projects can provide useful insights at the organization level. For example, analysis of the time-per-level and the most frequent development paths/cycles can bring to light operational bottlenecks.
Compared to conventional software engineering metrics based on sprint stories and tickets, or time-tracking tools, MLTRL provides a more accurate analysis of ML workflows.

\subsection*{Communication and explanation}
A distinct advantage of MLTRL in practice is the nomenclature: an agreed upon grading scheme for the maturity of an AI technology, and a framework for how/when that technology fits within a product or system, enables everyone to communicate effectively and transparently.
MLTRL also acts as a gate for interpretability and explainability--at the granularity of individual models and algorithms, and more crucially from a holistic, systems standpoint. Notably the DARPA XAI\footnote{\href{https://www.darpa.mil/program/explainable-artificial-intelligence}{DARPA Explainable Artificial Intelligence (XAI)}}
program advocates for this advance in developing AI technologies; they suggest interpretability and explainability are necessary at various locations in an AI system to be sufficient for deployment as an AI product, otherwise leading to issues with ethics and bias. 


\subsection*{Robustness via uncertainty-aware ML}
How to design a reliable system from unreliable components has been a guiding question in the fields of computing and intelligence \cite{Neumann1956ProbabilisticLA}. In the case of AI/ML systems, we aim to build reliable systems with myriad unreliable components: noisy and faulty sensors, human and AI error, and so on. There is thus significant value to quantifying the myriad uncertainties, propagating them throughout a system, and arriving at a notion or measure of reliability. For this reason, although MLTRL applies generally to AI/ML methods and systems, we advocate for methods in the class of probabilistic ML, which naturally represent and manipulate uncertainty about models and predictions\cite{Ghahramani2015ProbabilisticML}.
These are Bayesian methods that use probabilities to represent \textit{aleatoric uncertainty}, measuring the noise inherent in the observations, and \textit{epistemic uncertainty}, accounting for uncertainty in the model itself (i.e., capturing our ignorance about which model generated the data).
In the simplest case, an uncertainty aware ML pipeline should quantify uncertainty at the points of sensor inputs or perception, prediction or model output, and decision or end-user action -- McAllister et al.\cite{McAllister2017ConcretePF} suggest this with Bayesian deep learning models for safer autonomous vehicle pipelines. We can achieve this sufficiently well in practice for simple systems. However, we do not yet have a principled, theoretically grounded, and generalizable way of propagating errors and uncertainties downstream and throughout more complex AI systems -- i.e., how to integrate different software, hardware, data, and human components while considering how errors and uncertainties propagate through the system. This is an important direction of our future work.


\hfill \break

\bibliographystyle{unsrt}
\bibliography{mltrl}

\begin{thebibliography}{10}

\bibitem{Henderson2018DeepRL}
Peter Henderson, Riashat Islam, Philip Bachman, Joelle Pineau, Doina Precup,
  and David Meger.
\newblock Deep reinforcement learning that matters.
\newblock In {\em AAAI}, 2018.

\bibitem{deeptech}
Arnaud de~la Tour, Massimo Portincaso, Kyle Blank, and Nicolas Goeldel.
\newblock The dawn of the deep tech ecosystem.
\newblock Technical report, The Boston Consulting Group, 2019.

\bibitem{Nasa2003NASASE}
NASA.
\newblock The {NASA} systems engineering handbook.
\newblock 2003.

\bibitem{dod}
United States~Department of~Defense.
\newblock Defense acquisition guidebook.
\newblock Technical report, U.S. Dept. of Defense, 2004.

\bibitem{Leslie2019UnderstandingAI}
D.~Leslie.
\newblock Understanding artificial intelligence ethics and safety.
\newblock {\em ArXiv}, abs/1906.05684, 2019.

\bibitem{mlworkflow}
Google.
\newblock Machine learning workflow.
\newblock
  \url{https://cloud.google.com/mlengine/docs/tensorflow/ml-solutions-overview}.
\newblock Accessed: 2020-12-13.

\bibitem{Lavin2020TechnologyRL}
Alexander Lavin and Gregory Renard.
\newblock Technology readiness levels for {AI} \& {ML}.
\newblock {\em ICML Workshop on Challenges Deploying ML Systems}, 2020.

\bibitem{Dasu2003ExploratoryDM}
T.~Dasu and T.~Johnson.
\newblock Exploratory data mining and data cleaning.
\newblock 2003.

\bibitem{Janssen2020DataGO}
M.~Janssen, P.~Brous, Elsa Estevez, L.~Barbosa, and T.~Janowski.
\newblock Data governance: Organizing data for trustworthy artificial
  intelligence.
\newblock {\em Gov. Inf. Q.}, 37:101493, 2020.

\bibitem{Shahriari2016TakingTH}
B.~Shahriari, Kevin Swersky, Ziyu Wang, R.~Adams, and N.~D. Freitas.
\newblock Taking the human out of the loop: A review of bayesian optimization.
\newblock {\em Proceedings of the IEEE}, 104:148--175, 2016.

\bibitem{Ramakrishnan2020TowardsCD}
Goutham Ramakrishnan, A.~Nori, Hannah Murfet, and Pashmina Cameron.
\newblock Towards compliant data management systems for healthcare ml.
\newblock {\em ArXiv}, abs/2011.07555, 2020.

\bibitem{Bhatt2020ExplainableML}
Umang Bhatt, Alice Xiang, S.~Sharma, Adrian Weller, Ankur Taly, Yunhan Jia,
  Joydeep Ghosh, Ruchir Puri, Jos{\'e} M.~F. Moura, and P.~Eckersley.
\newblock Explainable machine learning in deployment.
\newblock {\em Proceedings of the 2020 Conference on Fairness, Accountability,
  and Transparency}, 2020.

\bibitem{Li2020FederatedLC}
Tian Li, Anit~Kumar Sahu, Ameet Talwalkar, and V.~Smith.
\newblock Federated learning: Challenges, methods, and future directions.
\newblock {\em IEEE Signal Processing Magazine}, 37:50--60, 2020.

\bibitem{Ryffel2018AGF}
T.~Ryffel, Andrew Trask, M.~Dahl, Bobby Wagner, J.~Mancuso, D.~Rueckert, and
  J.~Passerat-Palmbach.
\newblock A generic framework for privacy preserving deep learning.
\newblock {\em ArXiv}, abs/1811.04017, 2018.

\bibitem{Madry2018TowardsDL}
A.~Madry, Aleksandar Makelov, Ludwig Schmidt, D.~Tsipras, and Adrian Vladu.
\newblock Towards deep learning models resistant to adversarial attacks.
\newblock {\em ArXiv}, abs/1706.06083, 2018.

\bibitem{Zhao2018GeneratingNA}
Zhengli Zhao, Dheeru Dua, and Sameer Singh.
\newblock Generating natural adversarial examples.
\newblock {\em ArXiv}, abs/1710.11342, 2018.

\bibitem{Ribeiro2020BeyondAB}
Marco~T{\'u}lio Ribeiro, Tongshuang Wu, Carlos Guestrin, and Sameer Singh.
\newblock Beyond accuracy: Behavioral testing of nlp models with checklist.
\newblock In {\em ACL}, 2020.

\bibitem{Xie2011TestingAV}
Xiaoyuan Xie, Joshua W.~K. Ho, C.~Murphy, G.~Kaiser, B.~Xu, and T.~Chen.
\newblock Testing and validating machine learning classifiers by metamorphic
  testing.
\newblock {\em The Journal of systems and software}, 84 4:544--558, 2011.

\bibitem{DAmour2020UnderspecificationPC}
Alexander D'Amour, K.~Heller, D.~Moldovan, Ben Adlam, B.~Alipanahi, Alex
  Beutel, C.~Chen, Jonathan Deaton, Jacob Eisenstein, M.~Hoffman, Farhad
  Hormozdiari, N.~Houlsby, Shaobo Hou, Ghassen Jerfel, Alan Karthikesalingam,
  M.~Lucic, Y.~Ma, Cory~Y. McLean, Diana Mincu, Akinori Mitani, A.~Montanari,
  Zachary Nado, V.~Natarajan, C.~Nielson, Thomas~F. Osborne, R.~Raman,
  K.~Ramasamy, Rory Sayres, J.~Schrouff, Martin Seneviratne, Shannon Sequeira,
  Harini Suresh, V.~Veitch, Max Vladymyrov, Xuezhi Wang, K.~Webster,
  S.~Yadlowsky, Taedong Yun, Xiaohua Zhai, and D.~Sculley.
\newblock Underspecification presents challenges for credibility in modern
  machine learning.
\newblock {\em ArXiv}, abs/2011.03395, 2020.

\bibitem{Breck2017TheMT}
Eric Breck, Shanqing Cai, E.~Nielsen, M.~Salib, and D.~Sculley.
\newblock The ml test score: A rubric for ml production readiness and technical
  debt reduction.
\newblock {\em 2017 IEEE International Conference on Big Data (Big Data)},
  pages 1123--1132, 2017.

\bibitem{Botchkarev2019ANT}
A.~Botchkarev.
\newblock A new typology design of performance metrics to measure errors in
  machine learning regression algorithms.
\newblock {\em Interdisciplinary Journal of Information, Knowledge, and
  Management}, 14:045--076, 2019.

\bibitem{Duijm2015RecommendationsOT}
N.~Duijm.
\newblock Recommendations on the use and design of risk matrices.
\newblock {\em Safety Science}, 76:21--31, 2015.

\bibitem{Naud2020ManifoldsFU}
Louise Naud and Alexander Lavin.
\newblock Manifolds for unsupervised visual anomaly detection.
\newblock {\em ArXiv}, abs/2006.11364, 2020.

\bibitem{Gebru2018DatasheetsFD}
Timnit Gebru, J.~Morgenstern, Briana Vecchione, Jennifer~Wortman Vaughan,
  H.~Wallach, Hal Daum{\'e}, and K.~Crawford.
\newblock Datasheets for datasets.
\newblock {\em ArXiv}, abs/1803.09010, 2018.

\bibitem{Hutchinson2021TowardsAF}
B.~Hutchinson, A.~Smart, A.~Hanna, Emily~L. Denton, Christina Greer, Oddur
  Kjartansson, P.~Barnes, and Margaret Mitchell.
\newblock Towards accountability for machine learning datasets: Practices from
  software engineering and infrastructure.
\newblock {\em Proceedings of the 2021 ACM Conference on Fairness,
  Accountability, and Transparency}, 2021.

\bibitem{Schulam2017ReliableDS}
P.~Schulam and S.~Saria.
\newblock Reliable decision support using counterfactual models.
\newblock In {\em NIPS 2017}, 2017.

\bibitem{2018TowardsTM}
Towards trustable machine learning.
\newblock {\em Nature Biomedical Engineering}, 2:709--710, 2018.

\bibitem{Ghahramani2015ProbabilisticML}
Zoubin Ghahramani.
\newblock Probabilistic machine learning and artificial intelligence.
\newblock {\em Nature}, 521:452--459, 2015.

\bibitem{McAllister2017ConcretePF}
Rowan McAllister, Yarin Gal, Alex Kendall, Mark van~der Wilk, A.~Shah,
  R.~Cipolla, and Adrian Weller.
\newblock Concrete problems for autonomous vehicle safety: Advantages of
  bayesian deep learning.
\newblock In {\em IJCAI}, 2017.

\bibitem{Roberts2021COVID}
Michael Roberts, Derek Driggs, Matthew Thorpe, Julian Gilbey, Michael Yeung,
  Stephan Ursprung, Angelica~I. Avil{\'e}s-Rivero, Christian Etmann, Cathal
  McCague, Lucian Beer, Jonathan~R. Weir-McCall, Zhongzhao Teng, Effrossyni
  Gkrania-Klotsas, James H.~F. Rudd, Evis Sala, and Carola-Bibiane
  Sch{\"o}nlieb.
\newblock Common pitfalls and recommendations for using machine learning to
  detect and prognosticate for covid-19 using chest radiographs and ct scans.
\newblock {\em Nature Machine Intelligence}, 3:199--217, 2021.

\bibitem{Tobin2017DomainRF}
J.~Tobin, Rachel~H Fong, Alex Ray, J.~Schneider, W.~Zaremba, and P.~Abbeel.
\newblock Domain randomization for transferring deep neural networks from
  simulation to the real world.
\newblock {\em 2017 IEEE/RSJ International Conference on Intelligent Robots and
  Systems (IROS)}, pages 23--30, 2017.

\bibitem{Juliani2018UnityAG}
Arthur Juliani, Vincent-Pierre Berges, Esh Vckay, Yuan Gao, Hunter Henry,
  M.~Mattar, and D.~Lange.
\newblock Unity: A general platform for intelligent agents.
\newblock {\em ArXiv}, abs/1809.02627, 2018.

\bibitem{Hinterstoier2019AnAS}
Stefan Hinterstoi{\ss}er, Olivier Pauly, Tim~Hauke Heibel, Martina Marek, and
  Martin Bokeloh.
\newblock An annotation saved is an annotation earned: Using fully synthetic
  training for object instance detection.
\newblock {\em ArXiv}, abs/1902.09967, 2019.

\bibitem{Perception2021}
Steve Borkman, Adam Crespi, Saurav Dhakad, Sujoy Ganguly, Jonathan Hogins,
  You{-}Cyuan Jhang, Mohsen Kamalzadeh, Bowen Li, Steven Leal, Pete Parisi,
  Cesar Romero, Wesley Smith, Alex Thaman, Samuel Warren, and Nupur Yadav.
\newblock Unity perception: Generate synthetic data for computer vision.
\newblock {\em CoRR}, abs/2107.04259, 2021.

\bibitem{Cranmer2020TheFO}
K.~Cranmer, J.~Brehmer, and Gilles Louppe.
\newblock The frontier of simulation-based inference.
\newblock {\em Proceedings of the National Academy of Sciences}, 117:30055 --
  30062, 2020.

\bibitem{Meent2018AnIT}
Jan-Willem van~de Meent, Brooks Paige, H.~Yang, and Frank Wood.
\newblock An introduction to probabilistic programming.
\newblock {\em ArXiv}, abs/1809.10756, 2018.

\bibitem{Baydin2019EtalumisBP}
Atilim~G{\"u}nes Baydin, Lei Shao, W.~Bhimji, L.~Heinrich, Lawrence Meadows,
  Jialin Liu, Andreas Munk, Saeid Naderiparizi, Bradley Gram-Hansen, Gilles
  Louppe, Mingfei Ma, X.~Zhao, P.~Torr, V.~Lee, K.~Cranmer, Prabhat, and
  F.~Wood.
\newblock Etalumis: bringing probabilistic programming to scientific simulators
  at scale.
\newblock {\em Proceedings of the International Conference for High Performance
  Computing, Networking, Storage and Analysis}, 2019.

\bibitem{Gleisberg2009EventGW}
T.~Gleisberg, S.~H{\"o}che, F.~Krauss, M.~Sch{\"o}nherr, S.~Schumann,
  F.~Siegert, and J.~Winter.
\newblock Event generation with sherpa 1.1.
\newblock {\em Journal of High Energy Physics}, 2009:007--007, 2009.

\bibitem{Blei2014BuildCC}
David~M. Blei.
\newblock Build, compute, critique, repeat: Data analysis with latent variable
  models.
\newblock 2014.

\bibitem{Amershi2019SoftwareEF}
Saleema Amershi, Andrew Begel, Christian Bird, Robert DeLine, Harald~C. Gall,
  Ece Kamar, Nachiappan Nagappan, Besmira Nushi, and Thomas Zimmermann.
\newblock Software engineering for machine learning: A case study.
\newblock {\em 2019 IEEE/ACM 41st International Conference on Software
  Engineering: Software Engineering in Practice (ICSE-SEIP)}, 2019.

\bibitem{Ambrosino1995TheUO}
R.~Ambrosino, B.~Buchanan, G.~Cooper, and Marvin~J. Fine.
\newblock The use of misclassification costs to learn rule-based decision
  support models for cost-effective hospital admission strategies.
\newblock {\em Proceedings. Symposium on Computer Applications in Medical
  Care}, pages 304--8, 1995.

\bibitem{griffith2020collider}
Gareth~J Griffith, Tim~T Morris, Matthew~J Tudball, Annie Herbert, Giulia
  Mancano, Lindsey Pike, Gemma~C Sharp, Jonathan Sterne, Tom~M Palmer,
  George~Davey Smith, et~al.
\newblock Collider bias undermines our understanding of covid-19 disease risk
  and severity.
\newblock {\em Nature communications}, 11(1):1--12, 2020.

\bibitem{Pearl2018TheoreticalIT}
J.~Pearl.
\newblock Theoretical impediments to machine learning with seven sparks from
  the causal revolution.
\newblock {\em Proceedings of the Eleventh ACM International Conference on Web
  Search and Data Mining}, 2018.

\bibitem{Nguyen2017DoubleadjustmentIP}
T.~Nguyen, G.~Collins, J.~Spence, J.~Daur{\`e}s, P.~Devereaux, P.~Landais, and
  Y.~Le Manach.
\newblock Double-adjustment in propensity score matching analysis: choosing a
  threshold for considering residual imbalance.
\newblock {\em BMC Medical Research Methodology}, 17, 2017.

\bibitem{Eckles2017BiasAH}
D.~Eckles and E.~Bakshy.
\newblock Bias and high-dimensional adjustment in observational studies of peer
  effects.
\newblock {\em ArXiv}, abs/1706.04692, 2017.

\bibitem{Xu2020SplitTreatmentAT}
Yanbo Xu, Divyat Mahajan, Liz Manrao, A.~Sharma, and E.~Kiciman.
\newblock Split-treatment analysis to rank heterogeneous causal effects for
  prospective interventions.
\newblock {\em ArXiv}, abs/2011.05877, 2020.

\bibitem{Richens2020ImprovingTA}
Jonathan~G Richens, C.~M. Lee, and Saurabh Johri.
\newblock Improving the accuracy of medical diagnosis with causal machine
  learning.
\newblock {\em Nature Communications}, 11, 2020.

\bibitem{Paleyes2020ChallengesID}
Andrei Paleyes, Raoul-Gabriel Urma, and N.~Lawrence.
\newblock Challenges in deploying machine learning: a survey of case studies.
\newblock {\em ArXiv}, abs/2011.09926, 2020.

\bibitem{Chernozhukov2018DoubleDebiasedML}
V.~Chernozhukov, D.~Chetverikov, M.~Demirer, E.~Duflo, Christian~L. Hansen,
  Whitney~K. Newey, and J.~Robins.
\newblock Double/debiased machine learning for treatment and structural
  parameters.
\newblock {\em Econometrics: Econometric \& Statistical Methods - Special
  Topics eJournal}, 2018.

\bibitem{veitch2020sense}
Victor Veitch and Anisha Zaveri.
\newblock Sense and sensitivity analysis: Simple post-hoc analysis of bias due
  to unobserved confounding.
\newblock {\em NeurIPS 2020, arXiv preprint arXiv:2003.01747}, 2020.

\bibitem{JENNISKENS201140}
P.~Jenniskens, P.S. Gural, L.~Dynneson, B.J. Grigsby, K.E. Newman, M.~Borden,
  M.~Koop, and D.~Holman.
\newblock Cams: Cameras for allsky meteor surveillance to establish minor
  meteor showers.
\newblock {\em Icarus}, 216(1):40 -- 61, 2011.

\bibitem{Ganju2020LearningsFF}
Siddha Ganju, Anirudh Koul, Alexander Lavin, J.~Veitch-Michaelis, Meher Kasam,
  and J.~Parr.
\newblock Learnings from frontier development lab and spaceml - ai accelerators
  for nasa and esa.
\newblock {\em ArXiv}, abs/2011.04776, 2020.

\bibitem{Zoghbi2017SearchingFL}
S.~Zoghbi, M.~Cicco, A.~P. Stapper, A.~J. Ordonez, J.~Collison, P.~S. Gural,
  S.~Ganju, J.-L. Galache, and P.~Jenniskens.
\newblock Searching for long-period comets with deep learning tools.
\newblock In {\em Deep Learning for Physical Science Workshop, NeurIPS}, 2017.

\bibitem{JENNISKENS201821}
Peter Jenniskens, Jack Baggaley, Ian Crumpton, Peter Aldous, Petr Pokorny,
  Diego Janches, Peter~S. Gural, Dave Samuels, Jim Albers, Andreas Howell, Carl
  Johannink, Martin Breukers, Mohammad Odeh, Nicholas Moskovitz, Jack Collison,
  and Siddha Ganju.
\newblock A survey of southern hemisphere meteor showers.
\newblock {\em Planetary and Space Science}, 154:21 -- 29, 2018.

\bibitem{Cohn1994ActiveLW}
D.~Cohn, Zoubin Ghahramani, and Michael~I. Jordan.
\newblock Active learning with statistical models.
\newblock In {\em NIPS}, 1994.

\bibitem{Gal2017DeepBA}
Y.~Gal, Riashat Islam, and Zoubin Ghahramani.
\newblock Deep bayesian active learning with image data.
\newblock {\em ArXiv}, abs/1703.02910, 2017.

\bibitem{Sculley2015HiddenTD}
D.~Sculley, Gary Holt, Daniel Golovin, Eugene Davydov, Todd Phillips, Dietmar
  Ebner, Vinay Chaudhary, Michael Young, Jean-François Crespo, and Dan
  Dennison.
\newblock Hidden technical debt in machine learning systems.
\newblock In {\em NIPS}, 2015.

\bibitem{Abrahamsson2017AgileSD}
P.~Abrahamsson, Outi Salo, Jussi Ronkainen, and Juhani Warsta.
\newblock Agile software development methods: Review and analysis.
\newblock {\em ArXiv}, abs/1709.08439, 2017.

\bibitem{Kuhrmann2017HybridSA}
Marco Kuhrmann, Philipp Diebold, J{\"u}rgen M{\"u}nch, Paolo Tell, Vahid
  Garousi, Michael Felderer, Kitija Trektere, Fergal McCaffery, Oliver Linssen,
  Eckhart Hanser, and Christian~R. Prause.
\newblock Hybrid software and system development in practice: waterfall, scrum,
  and beyond.
\newblock {\em Proceedings of the 2017 International Conference on Software and
  System Process}, 2017.

\bibitem{Gelman2020Bayesian}
Andrew Gelman, Aki Vehtari, Daniel Simpson, Charles Margossian, Bob Carpenter,
  Yuling Yao, Lauren Kennedy, Jonah Gabry, Paul-Christian Burkner, and Martin
  Modrak.
\newblock Bayesian workflow.
\newblock {\em ArXiv}, abs/2011.01808, 2020.

\bibitem{Chapman2000CRISPDM1S}
P.~Chapman, J.~Clinton, R.~Kerber, T.~Khabaza, T.~Reinartz, C.~Shearer, and
  R.~Wirth.
\newblock Crisp-dm 1.0: Step-by-step data mining guide.
\newblock 2000.

\bibitem{Hohman2020UnderstandingAV}
Fred Hohman, Kanit Wongsuphasawat, Mary~Beth Kery, and Kayur Patel.
\newblock Understanding and visualizing data iteration in machine learning.
\newblock {\em Proceedings of the 2020 CHI Conference on Human Factors in
  Computing Systems}, 2020.

\bibitem{Amershi2014PowerTT}
Saleema Amershi, M.~Cakmak, W.~B. Knox, and T.~Kulesza.
\newblock Power to the people: The role of humans in interactive machine
  learning.
\newblock {\em AI Mag.}, 35:105--120, 2014.

\bibitem{Breck2019DataVF}
Eric Breck, Marty Zinkevich, Neoklis Polyzotis, Steven~Euijong Whang, and Sudip
  Roy.
\newblock Data validation for machine learning.
\newblock 2019.

\bibitem{Kumar2019FailureMI}
R.~Kumar, David~R. O'Brien, Kendra Albert, Salom{\'e} Vilj{\"o}en, and Jeffrey
  Snover.
\newblock Failure modes in machine learning systems.
\newblock {\em ArXiv}, abs/1911.11034, 2019.

\bibitem{Raji2020ClosingTA}
Inioluwa~Deborah Raji, Andrew Smart, Rebecca White, M.~Mitchell, Timnit Gebru,
  B.~Hutchinson, Jamila Smith-Loud, Daniel Theron, and P.~Barnes.
\newblock Closing the ai accountability gap: defining an end-to-end framework
  for internal algorithmic auditing.
\newblock {\em Proceedings of the 2020 Conference on Fairness, Accountability,
  and Transparency}, 2020.

\bibitem{Miksad2018HarnessingTP}
R.~Miksad and A.~Abernethy.
\newblock Harnessing the power of real‐world evidence (rwe): A checklist to
  ensure regulatory‐grade data quality.
\newblock {\em Clinical Pharmacology and Therapeutics}, 103:202 -- 205, 2018.

\bibitem{Larson2020RegulatoryFF}
D.~B. Larson, Hugh Harvey, D.~Rubin, Neville Irani, J.~R. Tse, and C.~Langlotz.
\newblock Regulatory frameworks for development and evaluation of artificial
  intelligence–based diagnostic imaging algorithms: Summary and
  recommendations.
\newblock {\em Journal of the American College of Radiology}, 2020.

\bibitem{Mehrabi2019ASO}
Ninareh Mehrabi, Fred Morstatter, N.~Saxena, Kristina Lerman, and A.~Galstyan.
\newblock A survey on bias and fairness in machine learning.
\newblock {\em ACM Computing Surveys (CSUR)}, 54:1 -- 35, 2019.

\bibitem{Ntoutsi2020BiasID}
Eirini Ntoutsi, P.~Fafalios, U.~Gadiraju, Vasileios Iosifidis, W.~Nejdl,
  Maria-Esther Vidal, S.~Ruggieri, F.~Turini, S.~Papadopoulos, Emmanouil
  Krasanakis, I.~Kompatsiaris, K.~Kinder-Kurlanda, Claudia Wagner, F.~Karimi,
  Miriam Fern{\'a}ndez, Harith Alani, B.~Berendt, Tina Kruegel, C.~Heinze,
  Klaus Broelemann, Gjergji Kasneci, T.~Tiropanis, and Steffen Staab.
\newblock Bias in data-driven ai systems - an introductory survey.
\newblock {\em ArXiv}, abs/2001.09762, 2020.

\bibitem{Jo2020LessonsFA}
E.~Jo and Timnit Gebru.
\newblock Lessons from archives: strategies for collecting sociocultural data
  in machine learning.
\newblock {\em Proceedings of the 2020 Conference on Fairness, Accountability,
  and Transparency}, 2020.

\bibitem{Wiens2020DiagnosingBI}
J.~Wiens, W.~Price, and M.~Sjoding.
\newblock Diagnosing bias in data-driven algorithms for healthcare.
\newblock {\em Nature Medicine}, 26:25--26, 2020.

\bibitem{Challen2019ArtificialIB}
R.~Challen, J.~Denny, M.~Pitt, L.~Gompels, T.~Edwards, and
  K.~Tsaneva-Atanasova.
\newblock Artificial intelligence, bias and clinical safety.
\newblock {\em BMJ Quality \& Safety}, 28:231 -- 237, 2019.

\bibitem{Obermeyer2019DissectingRB}
Z.~Obermeyer, B.~Powers, C.~Vogeli, and S.~Mullainathan.
\newblock Dissecting racial bias in an algorithm used to manage the health of
  populations.
\newblock {\em Science}, 366:447 -- 453, 2019.

\bibitem{Mitchell2019ModelCF}
Margaret Mitchell, Simone Wu, Andrew Zaldivar, Parker Barnes, Lucy Vasserman,
  Ben Hutchinson, Elena Spitzer, Inioluwa~Deborah Raji, and Timnit Gebru.
\newblock Model cards for model reporting.
\newblock {\em Proceedings of the Conference on Fairness, Accountability, and
  Transparency}, 2019.

\bibitem{Rivera2020GuidelinesFC}
Samantha~Cruz Rivera, Xiaoxuan Liu, A.~Chan, A.~K. Denniston, and M.~Calvert.
\newblock Guidelines for clinical trial protocols for interventions involving
  artificial intelligence: the spirit-ai extension.
\newblock {\em Nature Medicine}, 26:1351 -- 1363, 2020.

\bibitem{Szajnfarber2014ManagingII}
Z.~Szajnfarber.
\newblock Managing innovation in architecturally hierarchical systems: Three
  switchback mechanisms that impact practice.
\newblock {\em IEEE Transactions on Engineering Management}, 61:633--645, 2014.

\bibitem{Zhou2018ComparativeSO}
H.~Zhou and Y.~He.
\newblock Comparative study of okr and kpi.
\newblock {\em DEStech Transactions on Economics, Business and Management},
  2018.

\bibitem{Neumann1956ProbabilisticLA}
J.~Neumann.
\newblock Probabilistic logic and the synthesis of reliable organisms from
  unreliable components.
\newblock 1956.

\end{thebibliography}

\section*{Acknowledgements}
The authors would like to thank Gur Kimchi, Carl Henrik Ek and Neil Lawrence for valuable discussions about this project.

\section*{Author contributions statement}

A.L. conceived of the original ideas and framework, with significant contributions towards improving the framework from all co-authors. A.L. initiated the use of MLTRL in practice, including the neuropathology test case discussed here.
C.G-L. contributed insight regarding causal AI, including the section on counterfactual diagnosis. C.G-L. also made significant contributions broadly in the paper, notably in the Methods descriptions and paper revisions.
Si.G. contributed the spacecraft test case, along with early insights in the framework definitions.
A.V. contributed to the definition of later stages involving deployment (as did A.G.), and comparison with traditional software workflows.
Both E.X. and Y.G. provided insights regarding AI in academia, and Y.G. additionally contributed to the uncertainty quantification methods.
Su.G. and D.L. contributed the computer vision test case.
A.G.B. contributed the particle physics test case, and significant reviews of the writeup.
A.S. contributed insights related to causal ML and AI ethics.
D.N. provided valuable feedback on the overall framework, and contributed significantly with the details on ``switchback mechanisms''.
S.Z. contributed to multiple paper revisions, with emphasis on clarity and applicability to broad ML users and teams.
J.P. contributed to multiple paper revisions, and to deploying the systems ML methods broadly in practice for Earth and space sciences.
--same goes for C.M., with additional feedback overall on the methods.
All co-authors discussed the content and contributed to editing the manuscript.

\section*{Competing interests}
The authors declare no competing interests.

\section*{Additional information}
\textbf{Correspondence} and requests for materials should be addressed to A.L.

\end{document}